\documentclass[10pt]{article}
\usepackage{graphicx}
\usepackage{natbib}
\usepackage{amsmath}
\usepackage{amssymb}
\usepackage[small]{caption}
\usepackage{subfig}
\usepackage{geometry}
\usepackage{color}
\usepackage{algorithm}
\usepackage{algorithmic}
\newcommand{\prob}{\mathrm{Pr}}
\newcommand{\expect}{\mathbb{E}}

\newcommand{\normal}{{\cal N}}

\newcommand{\entropy}{\mathcal{H}}
\newcommand{\kldiv}{{\mathrm{D}_{\rm KL}}}
\newcommand{\klBars}{{\,\|\,}}
\newcommand{\given}{{\hspace{.1em}|\hspace{.1em}}}

\newcommand{\intD}{\textrm{d}}

\newcommand{\ndata}{{N}}

\newcommand{\nsamp}{{K}}
\newcommand{\ndist}{{T}}
\newcommand{\ncomp}{{K}}
\newcommand{\ndim}{{D}}

\newcommand{\dataIdx}{{i}}
\newcommand{\somethingI}[2]{{#1_{#2}}}

\newcommand{\sampleIdx}{{k}}
\newcommand{\somethingS}[2]{{#1^{(#2)}}}
\newcommand{\somethingIS}[3]{{#1_{#2}^{(#3)}}}
\newcommand{\distIdx}{{t}}

\newcommand{\nsStepIdx}{t}
\newcommand{\smallestSampleIdx}{\sampleIdx_\star}

\newcommand{\pmf}{{p}}
\newcommand{\pmfEstimate}{{\hat{p}}}
\newcommand{\pmfUnnorm}{{f}}
\newcommand{\pfn}{\mathcal{Z}}
\newcommand{\pmfProposal}{{q}}

\newcommand{\nsPrior}{\pi}
\newcommand{\nsConstrainedPrior}[1]{\breve{\nsPrior}_{#1}}
\newcommand{\nsLikelihood}{L}
\newcommand{\nsCutoff}{C}
\newcommand{\nsCutoffT}[1]{\nsCutoff_{#1}}
\newcommand{\nsVolume}{V}

\newcommand{\genState}{\mathbf{x}}
\newcommand{\genStateSpace}{\mathcal{X}}
\newcommand{\genStateS}[1]{{\somethingS{\genState}{#1}}}

\newcommand{\obs}{\mathbf{y}}
\newcommand{\obsI}[1]{{\somethingI{\obs}{#1}}}

\newcommand{\obsMat}{\mathbf{Y}}

\newcommand{\params}{{\boldsymbol \theta}}

\newcommand{\paramsS}[1]{{\somethingS{\params}{#1}}}

\newcommand{\latent}{\mathbf{z}}
\newcommand{\latentI}[1]{{\somethingI{\latent}{#1}}}
\newcommand{\latentS}[1]{{\somethingS{\latent}{#1}}}
\newcommand{\latentIS}[2]{{\somethingIS{\latent}{#1}{#2}}}

\newcommand{\model}{\mathcal{M}}
\newcommand{\data}{\mathcal{D}}

\newcommand{\weight}{{w}}
\newcommand{\weightS}[1]{{\somethingS{\weight}{#1}}}

\newcommand{\leftFactor}{\mathbf{U}}

\newcommand{\rightFactor}{\mathbf{V}}

\newcommand{\gComp}{\mathsf{G}}
\newcommand{\mComp}{\mathsf{M}}
\newcommand{\bComp}{\mathsf{B}}

\newcommand{\pfnEstimate}{{\hat \pfn}}

\newcommand{\pmfProposalForward}{{\pmfProposal_{for}}}
\newcommand{\pmfProposalBackward}{{\pmfProposal_{back}}}
\newcommand{\invTemp}{{\beta}}
\newcommand{\trans}{\mathcal{T}}
\newcommand{\transReverse}{\tilde{\mathcal{T}}}

\newcommand{\sumOfWeights}{{S}}

\newcommand{\latentPointEstimate}{{\latent^\star}}
\newcommand{\paramsPointEstimate}{{\params^\star}}
\newcommand{\leftFactorPointEstimate}{{\leftFactor^\star}}
\newcommand{\rightFactorPointEstimate}{{\rightFactor^\star}}

\newcommand{\pmfVariational}{{q}}
\newcommand{\variationalFunc}{\mathcal{F}}

\newcommand{\auxiliary}{\mathbf{v}}

\newcommand{\statGen}{{h}}

\newcommand{\pathParam}{{\beta}}

\bibpunct{(}{)}{;}{a}{,}{,}

\title{Sandwiching the marginal likelihood using \\ bidirectional Monte Carlo}
\author{Roger B.~Grosse \\  \small University of Toronto \\ \small rgrosse@cs.toronto.edu
\and
 Zoubin Ghahramani \\ \small University of Cambridge \\ \small zoubin@eng.cam.ac.uk
\and
Ryan P.~Adams \\ \small Twitter and Harvard University \\ \small rpa@seas.harvard.edu}
\date{}

\begin{document}
\maketitle

\begin{abstract}
Computing the marginal likelihood (ML) of a model requires marginalizing out all of the parameters and latent variables, a difficult high-dimensional summation or integration problem. To make matters worse, it is often hard to measure the accuracy of one's ML estimates. We present bidirectional Monte Carlo, a technique for obtaining accurate log-ML estimates on data simulated from a model. This method obtains stochastic lower bounds on the log-ML using annealed importance sampling or sequential Monte Carlo, and obtains stochastic upper bounds by running these same algorithms in reverse starting from an exact posterior sample. The true value can be sandwiched between these two stochastic bounds with high probability. Using the ground truth log-ML estimates obtained from our method, we quantitatively evaluate a wide variety of existing ML estimators on several latent variable models: clustering, a low rank approximation, and a binary attributes model. These experiments yield insights into how to accurately estimate marginal likelihoods.
\end{abstract}

\section{Introduction}

One commonly used model selection criterion is the marginal likelihood (ML) of the model, or $\pmf(\data \given \model_i)$, where $\data$ denotes the observed data and $\model_i$ denotes the model class \citep{kass-raftery-ml-review}. Marginal likelihood is an appealing criterion for several reasons. First, it can be plugged into Bayes' Rule to compute a posterior distribution over models, a practice known as Bayesian model comparison:
\[ \pmf(\model_i \given \data) = \frac{\pmf(\model_i)\, \pmf(\data \given \model_i)}{\sum_j \pmf(\model_j)\, \pmf(\data | \model_j)}. \]
Second, the ML criterion manages the tradeoff between model complexity and the goodness of fit to the data. Integrating out the model parameters results in a sophisticated form of Occam's Razor which penalizes the complexity of the model itself, rather than the specific parameterization \citep{bayesian-interpolation,bayesian-occams-razor}. Third, it is closely related to description length \citep{mdl}, a compression-based criterion for model selection. Finally, since the ML can be decomposed into a product of predictive likelihoods, it implicitly measures a model's ability to make predictions about novel examples. For these reasons, marginal likelihood is often the model selection criterion of choice when one is able to compute it efficiently. It is widely used to compare Gaussian process models \citep{gpml} and Bayesian network structures \citep{teyssier-koller}, 
where either closed-form solutions or accurate, tractable approximations are available.

The main difficulty in applying marginal likelihood is that it is intractable to compute for most models of interest. Computing $\pmf(\data \given \model_i)$ requires marginalizing out all of the parameters and latent variables of the model, an extremely high-dimensional summation or integration problem. It is equivalent to computing a partition function, a problem which is \#P-hard for graphical models in general. Much effort has been spent finding ways to compute or approximate partition functions for purposes of model selection \citep{kass-raftery-ml-review,bridge-sampling,gelman98,ais,ais-rbm,variational-bayes,chibs-method,chib-style,nested-sampling,evaluating-topic-models}. Different partition function estimation algorithms vary greatly in their accuracy, computational efficiency, and ease of implementation. Unfortunately, it is often difficult to determine which method to use in a given situation.

ML estimators can give inaccurate results in several ways. Subtle implementation bugs can lead to extremely inaccurate estimates with little indication that anything is amiss. Most estimators are based on sampling or optimization algorithms, and a failure to explore important modes of the posterior can also lead to highly inaccurate estimates. It is common for a particular algorithm to consistently under- or overestimate the true ML, even when the variance of the estimates appears to be small \citep[e.g.][]{neal-worst-monte-carlo}.

A major obstacle to developing effective ML estimators, and partition function estimators more generally, is that it is difficult even to know whether one's approximate ML estimates are accurate. The output of an ML estimator is a scalar value, and typically one does not have independent access to that value (otherwise one would not need to run the estimator). In a handful of cases, such as Ising models \citep{ising-poly-time}, one can tractably approximate the partition function to arbitrary accuracy using specialized algorithms. These models can be used to benchmark partition function estimators. However, such polynomial-time approximation schemes may not be known for any models sufficiently similar to the one whose ML needs to be measured. Alternatively, one can test the estimator on small problem instances for which the partition function can be easily computed \citep[e.g.][]{ais-rbm}, but this might not accurately reflect how it will perform on hard instances. It is also common to run multiple estimators and evaluate them based on their consistency with each other, under the implicit assumption that multiple algorithms are unlikely to fail in the same way. However, it has been observed that seemingly very different algorithms can report nearly the same value with high confidence, yet be very far from the true value (Iain Murray, personal communication).

In this paper, we present bidirectional Monte Carlo, a method for accurately estimating the ML for data \emph{simulated} from a model by sandwiching the true value between stochastic upper and lower bounds. In the limit of infinite computation, the two stochastic bounds are guaranteed to converge, so one need only run the algorithms for enough steps to achieve the desired accuracy. The technique is applicable both in the setting of ML estimation (where one integrates out both the parameters and the latent variables for an entire dataset) and in the setting of held-out likelihood estimation (where the parameters are fixed and one wants to integrate out the latent variables for a single data case).

As of yet, we do not know of a way to obtain accurate ML upper bounds on real-world data. However, the ability to accurately estimate ML on simulated data has several implications. First, one can measure the accuracy of an ML estimator on simulated data with similar characteristics to the real-world data. This can give a rough indication of how accurate the results would be in practice, even though there is no rigorous guarantee. Second, one can use this technique to construct a rigorous testbed for quantitatively evaluating ML estimators. This can be used to guide development of sampling algorithms, and perhaps even to optimize algorithm hyperparameters using Bayesian optimization \citep{bayesian-optimization}. Finally, this method also provides a rigorous way to quantitatively evaluate MCMC transition operators. In particular, the gap between the stochastic upper and lower bounds is itself a stochastic upper bound on the KL divergence of the distribution of approximate samples from the true posterior. By testing one's approximate posterior sampler on simulated data, one can get a rough idea of how it is likely to perform in practice, at least if one trusts that the simulated data are sufficiently realistic. 

The organization of this paper is as follows. Section~\ref{sec:background} provides background on existing ML estimators which this work builds on. In Section~\ref{sec:eval-ground-truth}, we describe bidirectional Monte Carlo, our method for sandwiching the marginal likelihood for simulated data. Section~\ref{sec:other-ml-estimators} describes some additional ML estimators which we compare in our experiments. Finally, in Section~\ref{sec:experiments}, we present experiments where we used our method to compute ML values on simulated data for three models: a clustering model, a low rank factorization, and a binary attribute model. We compare a wide variety of existing ML estimators on these models and conclude with recommendations for how the ML should be estimated in practice.

\section{Preliminaries}

Throughout this paper, in some cases it will be convenient to discuss the general partition function estimation problem, and in other cases it will be more convenient to discuss marginal likelihood estimation in particular. In the general partition function estimation setting, we have an unnormalized probability distribution $\pmfUnnorm$ over states $\genState \in \genStateSpace$, and we wish to evaluate the partition function $\pfn = \sum_{\genState \in \genStateSpace} \pmfUnnorm(\genState)$.

When we discuss marginal likelihood in particular, we let $\obs$ denote the observations, $\params$ denote the parameters, and $\latent$ denote the latent variables. (For instance, in a clustering model, $\params$ might correspond to the locations of the cluster centers and $\latent$ might correspond to the assignments of data points to clusters.) For notational convenience, we will assume continuous parameters and discrete latent variables, but this is not required by any of the algorithms we discuss. One defines the model by way of a joint distribution which we assume obeys the factorization
\begin{align}
\pmf(\params, \obs, \latent) &= \pmf(\params) \pmf(\latent \given \params) \pmf(\obs \given \latent, \params) \\
&= \pmf(\params) \prod_{\dataIdx=1}^\ndata \pmf(\latentI{\dataIdx} \given \params)\, \pmf(\obsI{\dataIdx} \given \latentI{\dataIdx}, \params)
\end{align}
We are interested in computing the marginal likelihood
\begin{equation}
 \pmf(\obs) = \int \pmf(\params) \prod_{\dataIdx=1}^\ndata \sum_{\latentI{\dataIdx}} \pmf(\latentI{\dataIdx} \given \params)\, \pmf(\obsI{\dataIdx} \given \latentI{\dataIdx}, \params)\, \mathrm{d}\params. \label{eqn:marginal-directed}
\end{equation}

Marginal likelihood estimation can be seen as a special case of partition function estimation, where $\genState = (\params, \latent)$, and $\pmfUnnorm$ is the joint distribution $\pmf(\params, \latent, \obs)$ viewed as a function of $\params$ and $\latent$.

\section{Background}
\label{sec:background}

Our method for sandwiching the marginal likelihood builds closely upon prior work on marginal likelihood estimation, in particular annealed importance sampling (AIS) and sequential Monte Carlo (SMC). This section reviews the techniques which we use in our procedure. Discussion of alternative ML estimators is deferred to Section~\ref{sec:other-ml-estimators}.

\subsection{Likelihood weighting}
\label{sec:likelihood-weighting}

Partition function estimators are often constructed from simple importance sampling (SIS). In particular, suppose we wish to compute the partition function $\pfn = \sum_\genState \pmfUnnorm(\genState)$. We generate a collection of samples $\genStateS{1}, \ldots, \genStateS{\nsamp}$ from a proposal distribution $\pmfProposal$ whose support contains the support of $\pmf$, and compute the estimate
\begin{align}
\pfnEstimate &= \frac{1}{\nsamp} \sum_{\sampleIdx=1}^\nsamp \weightS{\sampleIdx} \triangleq \frac{1}{\nsamp} \sum_{\sampleIdx=1}^\nsamp \frac{\pmfUnnorm(\genStateS{\sampleIdx})}{\pmfProposal(\genStateS{\sampleIdx})}. \label{eqn:sis-estimator}
\end{align}
This is an unbiased estimator of $\pfn$, because of the identity
\begin{align}
\expect_{\genState \sim \pmfProposal} \left[ \frac{\pmfUnnorm(\genState)}{\pmfProposal(\genState)} \right] &= \pfn. \label{eqn:sis-identity}
\end{align}

Likelihood weighting is a special case of this approach where the prior is used as the proposal distribution. In particular, for estimating the held-out likelihood of a directed model, latent variables $\latentS{1}, \ldots, \latentS{\nsamp}$ are sampled from $\pmf(\latent ; \params)$. By inspection, the weight $\weightS{\sampleIdx}$ is simply the data likelihood $\pmf(\obs \given \latentS{\sampleIdx} ; \params)$. This method can perform well if the latent space is small enough that the posterior can be adequately covered with a large enough number of samples. Unfortunately, likelihood weighting is unlikely to be an effective method for estimating marginal likelihood, because the model parameters would have to be sampled from the prior, and the chance that a random set of parameters happens to model the data well is vanishingly small.

\subsection{The harmonic mean of the likelihood}
\label{sec:harmonic-mean-estimator}

The harmonic mean estimator of \citet{harmonic-mean-estimator} is another estimator based on SIS. Here, the posterior is used as the proposal distribution, and the prior as the target distribution. By plugging these into Eqn.~\ref{eqn:sis-identity}, we obtain:
\begin{align}
\expect_{\params, \latent \sim \pmf(\params, \latent \given \obs)} \left[ \frac{\pmf(\params, \latent)}{\pmf(\params, \latent, \obs)} \right] = \frac{1}{\pmf(\obs)}.
\end{align}
This suggests the following estimator: draw samples $\{\paramsS{\sampleIdx}, \latentS{\sampleIdx}\}_{\sampleIdx=1}^{\nsamp}$ from the posterior $\pmf(\params, \latent \given \obs)$, and compute weights $\weightS{\sampleIdx} = \pmf(\paramsS{\sampleIdx}, \latentS{\sampleIdx}) / \pmf(\paramsS{\sampleIdx}, \latentS{\sampleIdx}, \obs) = 1/ \pmf(\obs \given \paramsS{\sampleIdx}, \latentS{\sampleIdx})$. The weights $\weightS{\sampleIdx}$ are unbiased estimators of the \emph{reciprocal} of the marginal likelihood. The ML estimate, then, is computed from the harmonic mean of the likelihood values:
\begin{align}
\pmfEstimate(\obs) = \frac{\nsamp}{\sum_{\sampleIdx=1}^\nsamp \weightS{\sampleIdx}} = \frac{\nsamp}{\sum_{\sampleIdx=1}^\nsamp 1/\pmf(\obs \given \paramsS{\sampleIdx}, \latentS{\sampleIdx})}.
\end{align}
While simple to implement, this estimator is unlikely to perform well in practice \citep{harmonic-mean-estimator,neal-worst-monte-carlo}.

\subsection{Annealed importance sampling}
\label{sec:background-ais}

The problem with both likelihood weighting and the harmonic mean estimator is that each one is based on a single importance sampling computation between two very dissimilar distributions. A more effective method is to bridge between the two distributions using a sequence of intermediate distributions. Annealed importance sampling \citep[AIS;][]{ais} is one algorithm based on this idea, and is widely used for estimating partition functions. Mathematically, the algorithm takes as input a sequence of $\ndist$ distributions $\pmf_1, \ldots, \pmf_\ndist$, with $\pmf_\distIdx(\genState) = \pmfUnnorm_\distIdx(\genState) / \pfn_\distIdx$, where $\pmf_\ndist$ is the target distribution and $\pmf_1$ is a tractable initial distribution, \emph{i.e.}~one for which we can efficiently evaluate the normalizing constant and generate exact samples. Most commonly, the intermediate distributions are taken to be geometric averages of the initial and target distributions: $\pmfUnnorm_\distIdx(\genState) = \pmfUnnorm_1(\genState)^{1-\invTemp_\distIdx}\pmfUnnorm_\ndist(\genState)^{\invTemp_\distIdx}$, where the $\invTemp_\distIdx$ are monotonically increasing parameters with $\invTemp_1 = 0$ and $\invTemp_\ndist = 1$.

The AIS procedure, shown in Algorithm \ref{alg:background-ais}, involves applying a sequence of MCMC transition operators $\trans_1, \ldots, \trans_\ndist$, where $\trans_\distIdx$ leaves $\pmf_\distIdx$ invariant. The result of the algorithm is a weight $\weight$ which is an unbiased estimator of the ratio of partition functions $\pfn_\ndist / \pfn_1$. Since $\pfn_1$ is typically known, $\pfn_1 \weight$ can be viewed as an unbiased estimator of $\pfn_\ndist = \pfn$.

\begin{algorithm}[t]
\begin{algorithmic}
	\FOR{$\sampleIdx = 1 \textrm{ to } \nsamp$} 
		\STATE $\genState_1 \gets$ sample from $\pmf_1(\genState)$
		\STATE $\weightS{\sampleIdx} \gets \pfn_1$
		\FOR{$\distIdx = 2 \textrm{ to } \ndist$} 
			\STATE $\weightS{\sampleIdx} \gets \weightS{\sampleIdx} \frac{\pmfUnnorm_{\distIdx}(\genState_{\distIdx-1})}{\pmfUnnorm_{\distIdx - 1}(\genState_{\distIdx-1})}$
			\STATE $\genState_\distIdx \gets$ sample from $\trans_\distIdx\left(\genState \, \middle| \, \genState_{\distIdx-1}\right)$
		\ENDFOR
	\ENDFOR
	\RETURN $\pfnEstimate = \sum_{\sampleIdx=1}^\nsamp \weightS{\sampleIdx}/\nsamp$
\end{algorithmic}
\caption{Annealed Importance Sampling}
\label{alg:background-ais}
\end{algorithm}

For purposes of evaluating marginal likelihood, $\pmfUnnorm_1$ is the prior distribution $\pmf(\params, \latent)$, and $\pmfUnnorm_\ndist$ is the joint distribution $\pmf(\params, \latent, \obs)$ (viewed as an unnormalized distribution over $\params$ and $\latent$). Because $\obs$ is fixed, the latter is proportional to the posterior $\pmf(\params, \latent \given \obs)$. The intermediate distributions are given by geometric averages of the prior and the posterior, which is equivalent to raising the likelihood term to a power less than 1:
\begin{align}
\pmfUnnorm_\distIdx(\params, \latent) = \pmf(\params, \latent)\, \pmf(\obs \given \params, \latent)^{\invTemp_\distIdx}.
\end{align}
Note that this form of annealing can destroy the directed factorization structure which is present in the prior and the joint distribution. Conditional independencies satisfied by the original model may not hold in the intermediate distributions. Unfortunately, this can make the implementation of MCMC operators for the intermediate distributions considerably more complicated compared to the analogous operators applied to the posterior. 

AIS can be justified as an instance of SIS over an extended state space \citep{ais}. In particular, the full set of states $\genState_1, \ldots, \genState_{\ndist-1}$ visited by the algorithm has a joint distribution (which we call the \emph{forward distribution}) represented by:
\begin{align}
\pmfProposalForward(\genState_1, \ldots, \genState_{\ndist-1}) &= \pmf_1(\genState_1)\, \trans_2(\genState_2 \given \genState_1) \cdots \trans_{\ndist-1}(\genState_{\ndist-1} \given \genState_{\ndist - 2}).
\end{align}
We can also postulate a reverse chain, where $\genState_{\ndist-1}$ is first sampled exactly from the distribution $\pmf_\ndist$, and the transition operators are applied in the reverse order. (Note that the reverse chain cannot be explicitly simulated in general, since it requires sampling from $\pmf_\ndist$.) The joint distribution is given by:
\begin{align}
\pmfProposalBackward(\genState_1, \ldots, \genState_{\ndist-1}) &= \pmf_\ndist(\genState_{\ndist - 1})\, \trans_{\ndist-1}(\genState_{\ndist-2} \given \genState_{\ndist-1}) \cdots \trans_2(\genState_1 \given \genState_2).
\end{align}
If $\pmfProposalForward$ is used as a proposal distribution for $\pmfProposalBackward$, the importance weights come out to:
\begin{align}
\frac{\pmfProposalBackward(\genState_1, \ldots, \genState_\ndist)}{\pmfProposalForward(\genState_1, \ldots, \genState_\ndist)} &= \frac{\pmf_\ndist(\genState_{\ndist-1})}{\pmf_1(\genState_1)} \frac{\trans_2(\genState_1 \given \genState_2)}{\trans_2(\genState_2 \given \genState_1)} \cdots \frac{\trans_{\ndist-1}(\genState_{\ndist-2} \given \genState_{\ndist-1})}{\trans_{\ndist-1}(\genState_{\ndist-1} \given \genState_{\ndist-2})} \\
&= \frac{\pmf_\ndist(\genState_{\ndist-1})}{\pmf_1(\genState_1)} \frac{\pmfUnnorm_2(\genState_1)}{\pmfUnnorm_2(\genState_2)} \cdots \frac{\pmfUnnorm_{\ndist-1}(\genState_{\ndist-2})}{\pmfUnnorm_{\ndist-1}(\genState_{\ndist-1})} \label{eqn:ais-reversibility-step} \\
&= \frac{\pfn_1}{\pfn_\ndist} \weight,
\end{align}
where $\weight$ is the weight computed in Algorithm~\ref{alg:background-ais} and (\ref{eqn:ais-reversibility-step}) follows from the reversibility of $\trans_\distIdx$. Since this quantity corresponds to an importance weight between normalized distributions, its expectation is 1, and therefore $\expect[\weight] = \pfn_\ndist / \pfn_1$. 

Note also that $\pmfProposalBackward(\genState_{\ndist-1}) = \pmf_\ndist(\genState_{\ndist-1})$. Therefore, AIS can also be used as an importance sampler for $\pmf_\ndist$:
\begin{align}
\expect_{\pmfProposalForward}[\weight \statGen(\genState_{\ndist-1})] &= \frac{\pfn_\ndist}{\pfn_1} \expect_{\pmf_\ndist}[\statGen(\genState)]
\end{align}
for any statistic $\statGen$. (The partition function estimator corresponds to the special case where $\statGen(\genState) = 1$.) Ordinarily, one uses the normalized importance weights when estimating expectations.

%\TODO{variance approaches zero}

\subsection{Sequential Monte Carlo}
\label{sec:particle-filter}

Observe that the marginal distribution $\pmf(\obs)$ can be decomposed into a series of predictive distributions:
\begin{align}
\pmf(\obsI{1:\ndata}) &= \pmf(\obsI{1})\, \pmf(\obsI{2} \given \obsI{1}) \cdots \pmf(\obsI{\ndata} \given \obsI{1:\ndata-1}). 
\end{align}
(In this section, we use Matlab-style slicing notation.) Since the predictive likelihood terms can't be computed exactly, approximations are required. Sequential Monte Carlo (SMC) methods \citep{smc} use particles to represent the parameters and/or the latent variables. In each step, as a new data point is observed, the particles are updated to take into account the new information. While SMC is most closely associated with filtering problems where there are explicit temporal dynamics, it has also been successfully applied to models with no inherent temporal structure, such as the ones considered in this work. This is the setting that we focus on here.

SMC is a very broad family of algorithms, so we cannot summarize all of the advances. Instead, we give a generic implementation in Algorithm \ref{alg:particle-learning} where several decisions are left unspecified. In each step, the latent variables are sampled according to a proposal distribution $\pmfProposal$, which may optionally take into account the current data point. (Some examples are given below.) The weights are then updated according to the evidence, and the model parameters (and possibly latent variables) are updated based on the new evidence.

This procedure corresponds the most closely to the particle learning approach of \citet{particle-learning}, where $\latent$ is approximated in the typical particle filter framework, and $\params$ is resampled from the posterior after each update. Our formulation is slightly more general: since it may not be possible to sample $\params$ exactly from the posterior, we allow any MCMC operator to be used which preserves the posterior distribution. Furthermore, we allow $\latent$ to be included in the MCMC step as well. \citet{particle-learning} do not allow this, because it would require revisiting all of the data after every sampling step. However, we consider it because it may be advantageous to pay the extra cost in the interest of more accurate results.

Algorithm \ref{alg:particle-learning} leaves open the choice of $\pmfProposal(\latentI{\dataIdx} \given \obsI{\dataIdx}, \paramsS{\sampleIdx})$, the proposal distribution for the latent variables at the subsequent time step. The simplest method is to ignore the observations and sample $\latentIS{\dataIdx}{\sampleIdx}$ from the predictive distribution, \emph{i.e.}
\begin{align}
\pmfProposal(\latentI{\dataIdx} \given \obsI{\dataIdx}, \paramsS{\sampleIdx}) &= \pmf(\latentI{\dataIdx} \given \paramsS{\sampleIdx}). \label{eqn:particle-filter-proposal}
\end{align}
The particles are then weighted according to the observation likelihood:
\begin{align}
\weightS{\sampleIdx} &\gets \weightS{\sampleIdx} \pmf(\obsI{\dataIdx} \given \latentIS{\dataIdx}{\sampleIdx}, \paramsS{\sampleIdx}).
\end{align}
A more accurate method, used in the posterior particle filter,
%\TODO{cite},
is to sample $\latentIS{\dataIdx}{\sampleIdx}$ from the posterior:
\begin{align}
\pmfProposal(\latentI{\dataIdx} \given \obsI{\dataIdx}, \paramsS{\sampleIdx}) &= \pmf(\latentI{\dataIdx} \given \obsI{\dataIdx}, \paramsS{\sampleIdx}). \label{eqn:posterior-particle-filter-proposal}
\end{align}
In this case, the weight update corresponds to the predictive likelihood:
\begin{align}
\weightS{\sampleIdx} &\gets \weightS{\sampleIdx} \pmf(\obsI{\dataIdx} \given \paramsS{\sampleIdx}) \\
&= \weightS{\sampleIdx} \sum_{\latentI{\dataIdx}} \pmf(\latentI{\dataIdx} \given \paramsS{\sampleIdx}) \pmf(\obsI{\dataIdx} \given \latentI{\dataIdx}, \paramsS{\sampleIdx}) \label{eqn:posterior-particle-filter-update}.
\end{align}
Posterior particle filtering can result in considerably lower variance of the weights compared to standard particle filtering, and therefore better marginal likelihood estimates. (We note that the posterior particle filter can only be applied to those models for which posterior inference of latent variables is tractable.)

\begin{algorithm}[t]
\begin{algorithmic}
\begin{small}
	\FOR{$\sampleIdx = 1 \textrm{ to } \nsamp$}
		\STATE $\paramsS{\sampleIdx} \gets$ sample from $\pmf(\params)$
		\STATE $\weightS{\sampleIdx} \gets 1$
	\ENDFOR
	\FOR{$\dataIdx = 1 \textrm{ to } \ndist$}
		\FOR{$\sampleIdx = 1 \textrm{ to } \nsamp$}
			\STATE $\latentIS{\dataIdx}{\sampleIdx} \gets$ sample from $\pmfProposal(\latentI{\dataIdx} \given \obsI{\dataIdx}, \paramsS{\sampleIdx})$
			\STATE $\weightS{\sampleIdx} \gets \weightS{\sampleIdx} \pmf(\latentIS{\dataIdx}{\sampleIdx} \given \params) \pmf(\obsI{\dataIdx} \given \latentIS{\dataIdx}{\sampleIdx}, \paramsS{\sampleIdx}) / \pmfProposal(\latentI{\dataIdx} \given \obsI{\dataIdx}, \paramsS{\sampleIdx})$
		%\STATE $\weightS{\sampleIdx} \gets \weightS{\sampleIdx} \pmf(\obsI{\dataIdx} \given \latentI{\dataIdx}, \params)$
			\STATE $(\latentIS{1:\dataIdx}{\sampleIdx}, \paramsS{\sampleIdx}) \gets$ MCMC transition which leaves $\pmf(\latentI{1:\dataIdx}, \params \given \obsI{1:\dataIdx})$ invariant
		\ENDFOR
		\IF{resampling criterion met}
			\STATE Sample $(\latentIS{1:\dataIdx}{\sampleIdx}, \paramsS{\sampleIdx})$ proportionally to $\weightS{\sampleIdx}$
			\STATE $\sumOfWeights \gets \sum_{\sampleIdx=1}^\nsamp \weightS{\sampleIdx}$
			\FOR{$\sampleIdx = 1 \textrm{ to } \nsamp$}
				\STATE $\weightS{\sampleIdx} \gets \sumOfWeights / \nsamp$
			\ENDFOR
		\ENDIF
	\ENDFOR
	\RETURN $\pfnEstimate = \frac{1}{\nsamp}\sum_{\sampleIdx=1}^\nsamp \weightS{\sampleIdx}$
\end{small}
\end{algorithmic}
\caption{Particle learning}
\label{alg:particle-learning}
\end{algorithm}

For simplicity of notation, Algorithm \ref{alg:particle-learning} explicitly samples the model parameters $\params$. However, for models where $\params$ has a simple closed form depending on certain sufficient statistics of $\obs$ and $\latent$, it can be collapsed out analytically, giving a Rao-Blackwellized particle filter. The algorithm is the same as Algorithm \ref{alg:particle-learning}, except that steps involving $\params$ are ignored and the updates for $\latent$ and $\weight$ are modified:
\begin{align}
\latentIS{\dataIdx}{\sampleIdx} &\gets \textrm{sample from } \pmfProposal(\latentIS{\dataIdx}{\sampleIdx} \given \obsI{1:\dataIdx}, \latentIS{1:\dataIdx-1}{\sampleIdx}) \\
\weightS{\sampleIdx} &\gets \weightS{\sampleIdx}  \frac{\pmf(\latentIS{\dataIdx}{\sampleIdx} \given \latentIS{1:\dataIdx-1}{\sampleIdx})\, \pmf(\obsI{\dataIdx} \given \latentIS{1:\dataIdx}{\sampleIdx}, \obsI{1:\dataIdx-1})}{\pmfProposal(\latentIS{\dataIdx}{\sampleIdx} \given \obsI{1:\dataIdx}, \latentIS{1:\dataIdx-1}{\sampleIdx})}
\end{align}

\subsubsection{Relationship with AIS}
\label{sec:relationship-smc-ais}

While SMC is based on a different intuition from AIS, the underlying mathematics is equivalent. In particular, we discuss the unifying view of \citet{smc}. For simplicity, assume there is only a single particle, \emph{i.e.}~$\nsamp = 1$. While Algorithm \ref{alg:particle-learning} incrementally builds up the latent representation one data point at a time, we can imagine that all of the latent variables are explicitly represented at every step. Recall that AIS was defined in terms of a sequence of unnormalized distributions $\pmfUnnorm_\distIdx$ and MCMC transition operators $\trans_\distIdx$ which leave each distribution invariant. In this section, $\distIdx$ ranges from $0$ to $\ndist$, rather than $1$ to $\ndist$ as in Section \ref{sec:background-ais}.

\begin{figure}
\begin{center}
\includegraphics[width=0.5 \textwidth]{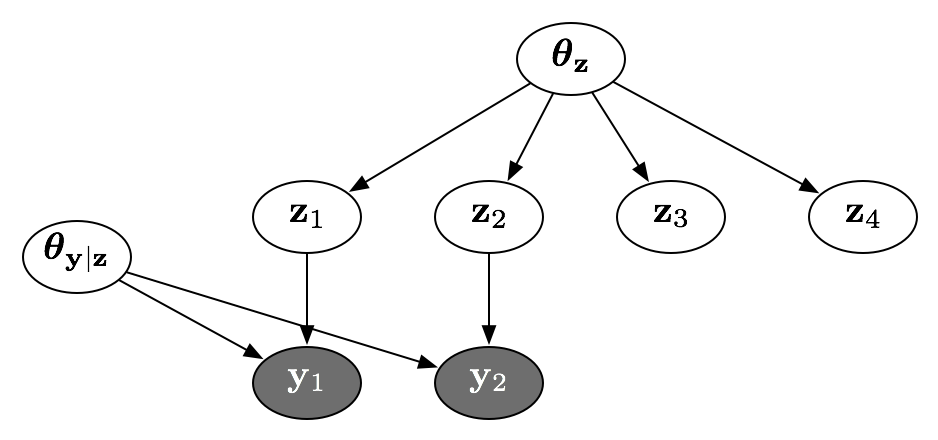}
\end{center}
\caption{The intermediate distribution $\pmfUnnorm_2(\obs, \latent) = \pmf(\params)\, \pmf(\latent \given \obs)\, \pmf(\obsI{1:2}\given\params, \latent)$. }
\label{fig:smc-intermediate}
\end{figure}

The intermediate distributions are constructed by including only a subset of the data likelihood terms:
\begin{align}
\pmfUnnorm_\distIdx(\params, \latent) &= \pmf(\params) \prod_{\dataIdx = 1}^\ndata \pmf(\latentI{\dataIdx}) \prod_{\dataIdx=1}^\distIdx \pmf(\obsI{\dataIdx} \given \params, \latentI{\dataIdx}).
\end{align}
This distribution is shown in Figure \ref{fig:smc-intermediate}. Since each distribution in the sequence differs from its predecessor simply by adding an additional observation likelihood term,
\begin{align}
\frac{\pmfUnnorm_\distIdx(\params, \latent)}{\pmfUnnorm_{\distIdx-1}(\params, \latent)} &= \pmf(\obsI{\distIdx} \given \params, \latentI{\distIdx}).
\end{align}
The transition operator first samples $\params$ from the conditional distribution $\pmf(\params \given \obsI{1:\distIdx}, \latentI{1:\distIdx})$, and then resamples $\latentI{\distIdx+1:\ndata}$ from $\pmf(\latentI{\distIdx+1:\ndata} \given \params)$. 

\section{Sandwiching the marginal likelihood}
\label{sec:eval-ground-truth}

In this section, we first define our notion of stochastic upper and lower bounds. We then describe our techniques for obtaining accurate stochastic upper and lower bounds on the log marginal likelihood for simulated data; in combination, these bounds give precise estimates of the true value. Stochastic lower bounds can be computed using various existing methods, as described in Section~\ref{sec:background-pfn-principles}. Our main technical contribution is a set of techniques for obtaining accurate stochastic upper bounds. These techniques require an exact sample from the posterior distribution, which one can obtain for simulated data. As described in Sections~\ref{sec:reverse-ais} and \ref{sec:shme}, the key idea is to run AIS or SMC in reverse, starting from the exact posterior sample. Since our final log-ML estimates are obtained by running AIS or SMC both forwards and backwards, we call our approach bidirectional Monte Carlo (BDMC).\footnote{While we limit our discussion to exact samples obtained from simulated data, other techniques have also been proposed for obtaining exact samples. For instance, this can be done for Ising models using coupling from the past \citep{coupling-from-the-past}. BDMC could be used in conjunction with such techniques, without the requirement of simulated data.}

\subsection{Obtaining stochastic lower bounds}
\label{sec:background-pfn-principles}

One can obtain stochastic lower bounds on the log-ML using a variety of existing algorithms, as we now outline. In this section, it is more convenient to discuss the general setting of partition function estimation.

Many partition function estimators, such as SIS (Section~\ref{sec:likelihood-weighting}) and AIS (Section~\ref{sec:background-ais}), are unbiased, \emph{i.e.}~$\expect[\pfnEstimate] = \pfn$.\footnote{In this context, unbiasedness can be misleading: because partition function estimates can vary over many orders of magnitude, it's common for an unbiased estimator to drastically underestimate $\pfn$ with overwhelming probability, yet occasionally return extremely large estimates. (An extreme example is likelihood weighting (Section \ref{sec:likelihood-weighting}), which is unbiased, but is extremely unlikely to give an accurate answer for a high-dimensional model.) Unless the estimator is chosen very carefully, the variance is likely to be extremely large, or even infinite.}
Since $\pfn$ can vary over many orders of magnitude, it is often more meaningful to talk about estimating $\log \pfn$, rather than $\pfn$. Unfortunately, unbiased estimators of $\pfn$ may correspond to biased estimators of $\log \pfn$. In particular, they are stochastic lower bounds, in two senses. First, because ML estimators are nonnegative estimators of a nonnegative quantity, Markov's inequality implies that $\prob(\pfnEstimate > a \pfn) < 1/a$. By taking the log, we find that
\begin{align}
\prob(\log \pfnEstimate > \log \pfn + b) &< e^{-b}. \label{eqn:markov}
\end{align}
In other words, the estimator is exceedingly unlikely to overestimate $\log \pfn$ by more than a few nats. One can improve this tail bound by combining multiple independent samples \citep{gogate-lower-bound}, but in the context of log-ML estimation, an error of a few nats is insignificant.

The other sense in which $\log \pfnEstimate$ is a stochastic lower bound on $\log \pfn$ follows from Jensen's inequality:
\begin{align}
\expect[\log \pfnEstimate] \leq \log \expect[\pfnEstimate] &= \log \pfn. \label{eqn:jensen}
\end{align}
In general, we will use the term \emph{stochastic lower bound} to refer to an estimator which satisfies Eqns.~\ref{eqn:markov} and \ref{eqn:jensen}.

Of course, it is not enough to have a stochastic lower bound; we would also like the estimates to be close to the true value. Fortunately, AIS and SMC are both consistent, in that they converge to the correct value in the limit of infinite computation. (For AIS, this means adding more intermediate distributions \citep{ais}; for SMC, it means adding more particles \citep{smc}.) We note that it is also possible to (deterministically) lower bound the log-ML using variational Bayes (discussed in more detail in Section~\ref{sec:variational-bayes}). However, variational Bayes does not enjoy the same consistency guarantees as AIS and SMC.

\subsection{Obtaining stochastic upper bounds}

Heuristically speaking, we would expect good upper bounds to be harder to obtain than good lower bounds. For a lower bound on the log-ML, it suffices to exhibit regions of high posterior mass. For an upper bound, one would have to demonstrate the absence of any additional probability mass. Indeed, while variational upper bounds have been proposed \citep{tree-reweighted}, we aren't aware of any practical stochastic upper bounds which achieve comparable accuracy to AIS. But suppose we are given a hint in the form of an exact posterior sample. Here, we propose methods which make use of an exact posterior sample to give accurate stochastic upper bounds on the log-ML.

As discussed in Section \ref{sec:harmonic-mean-estimator}, the harmonic mean estimator (HME) is derived from an unbiased estimate of the \emph{reciprocal} of the ML. The arguments of Section~\ref{sec:background-pfn-principles} show that such unbiased estimates of the reciprocal correspond to stochastic upper bounds on the log-ML. Unfortunately, there are two problems with simply using the HME: first, if approximate posterior samples are used, the estimator is not a stochastic upper bound, and in fact can underestimate the log-ML if the sampler failed to find an important mode. Second, as pointed out by \citet{neal-worst-monte-carlo} and further confirmed in our experiments, even when exact samples are used, the bound can be extremely poor.

For simulated data, it is possible to work around both of these issues. For the issue of finding exact posterior samples, observe that there are two different ways to sample from the joint distribution $\pmf(\params, \latent, \obs)$ over parameters $\params$, latent variables $\latent$, and observations $\obs$:
On one hand, we can simulate from the model by first sampling $(\params, \latent)$ from $\pmf(\params, \latent)$, and then sampling $\obs$ from $\pmf(\obs \given \params, \latent)$. Alternatively, we can first sample $\obs$ from $\pmf(\obs)$, and then sample $(\params, \latent)$ from the posterior $\pmf(\params, \latent \given \obs)$. Since these two processes sample from the same joint distribution, the $(\params, \latent)$ generated during forward sampling is also an exact sample from the posterior $\pmf(\params, \latent \given \obs)$. (The Geweke test \citep{geweke-test} is based on the same identity.) In other words, for a simulated dataset, we have available a single exact sample from the posterior, namely the parameters and latent variables used to generate the data.\footnote{More posterior samples can be obtained by running an MCMC algorithm starting from the original one. However, the statistics would likely be correlated with those of the original sample. We used a single exact sample for each of our experiments.}

The other problem with the HME is that the bound can be extremely poor even when computed with exact samples. This happens because the estimator is based on simple importance sampling from the posterior to the prior --- two very dissimilar distributions in a high-dimensional space. As discussed in Section~\ref{sec:harmonic-mean-estimator}, the HME is essentially the mirror image of likelihood weighting, which is simple importance sampling from the prior to the posterior. Sections \ref{sec:background-ais} and \ref{sec:particle-filter} discussed two algorithms --- annealed importance sampling (AIS) and sequential Monte Carlo (SMC) --- which estimate marginal likelihoods though a series of much smaller importance sampling steps bridging from the prior to the posterior. The use of many small steps, rather than a single large step, typically results in far more accurate estimates of the log-ML. This suggests that, in place of the HME, one should use a series of small importance sampling steps bridging from the posterior to the prior. We now discuss two particular instantiations of this idea: reverse AIS and the sequential harmonic mean estimator.

\subsubsection{Reverse AIS}
\label{sec:reverse-ais}

In Section \ref{sec:background-ais}, we discussed an interpretation of AIS as simple importance sampling over an extended state space, where the proposal and target distributions correspond to forward and backward annealing chains. We noted that that the reverse chain generally could not be sampled from explicitly because it required an exact sample from $\pmf_\ndist(\genState)$ -- in this case, the posterior distribution. However, for simulated data, the reverse chain can be run starting from an exact sample as described above. The importance weights for the forward chain using the backward chain as a proposal distribution are given by:
\begin{align}
\frac{\pmfProposalForward(\genState_1, \ldots, \genState_\ndist)}{\pmfProposalBackward(\genState_1, \ldots, \genState_\ndist)} &= \frac{\pfn_\ndist}{\pfn_1} \weight, \label{eqn:ais-as-sis}
\end{align}
where
\begin{align}
\weight &\triangleq \frac{\pmfUnnorm_{\ndist-1}(\genState_{\ndist-1})}{\pmfUnnorm_\ndist(\genState_{\ndist-1})} \cdots \frac{\pmfUnnorm_1(\genState_1)}{\pmfUnnorm_2(\genState_1)}.
\end{align}
As in Section \ref{sec:background-ais}, because Eqn.~\ref{eqn:ais-as-sis} represents the importance weights between two normalized distributions, its expectation must be 1, and therefore $\expect[\weight] = \pfn_1 / \pfn_\ndist$. Since $\pmf_1$ is chosen to be the prior, $\pfn_1 = 1$, and we obtain the following estimate of the ML:
\begin{align}
\pmfEstimate_{back}(\obs) = \frac{\nsamp}{\sum_{\sampleIdx=1}^\nsamp \weightS{\sampleIdx}}. \label{eqn:reverse-ais-estimator}
\end{align}

This estimator corresponds to a stochastic upper bound on $\log \pmf(\obs)$, for reasons which mirror those given in Section~\ref{sec:background-pfn-principles}. By Markov's inequality, since $\expect[1/\pmfEstimate_{back}(\obs)] = 1/\pmf(\obs)$,
\begin{equation}
\prob \left(\log \pmfEstimate_{back}(\obs) < \log \pmf(\obs) - b \right) = \prob \left(\frac{1}{\pmfEstimate_{back}(\obs)} > \frac{e^{b}}{\pmf(\obs)} \right) < e^{-b}.
\end{equation}
Also, by Jensen's inequality,
\begin{equation}
\expect[\log \pmfEstimate_{back}(\obs)] = -\expect \left[\log \frac{1}{\pmfEstimate_{back}(\obs)} \right] \geq -\log \expect \left[ \frac{1}{\pmfEstimate_{back}(\obs)} \right] = -\log \frac{1}{\pmf(\obs)} = \log \pmf(\obs)
\end{equation}
Since this estimator is an instance of AIS, it inherits the consistency guarantees of AIS \citep{ais}.

\subsubsection{Sequential harmonic mean estimator}
\label{sec:shme}

It is also possible to run sequential Monte Carlo (SMC) in reverse to obtain a log-ML upper bound. We call the resulting algorithm the sequential harmonic mean estimator (SHME). Mathematically, this does not require any new ideas beyond those used in reverse AIS. As discussed in Section \ref{sec:relationship-smc-ais}, SMC with a single particle can be analyzed as a special case of AIS. Therefore, starting from an exact posterior sample, we can run the reverse chain for SMC as well. The resulting algorithm, which we call the sequential harmonic mean estimator (SHME), corresponds to starting with full observations and an exact posterior sample, and deleting one observation at a time. Each time an observation is deleted, the weights are updated with the likelihood of the observations, similarly to SMC. The difference is in how the weights are used: when the resampling criterion is met, the particles are sampled proportionally to the \emph{reciprocal} of their weights. Also, while SMC computes arithmetic means of the weights in the resampling step and in the final ML estimate, SHME uses the harmonic means of the weights.  

As in SMC, we leave open the choice of proposal distribution $\pmfProposal(\latentI{\dataIdx} \given \obsI{\dataIdx}, \paramsS{\sampleIdx})$. Possibilities include Eqns.~\ref{eqn:particle-filter-proposal} and \ref{eqn:posterior-particle-filter-proposal} of Section \ref{sec:particle-filter}. Unlike in standard SMC, the proposal distribution does not affect the sequence of states sampled in the algorithm---rather, it is used to update the weights. Proposal distributions which better match the posterior are likely to result in lower variance weights. 

\begin{algorithm}[t]
\begin{algorithmic}
\begin{small}
	\FOR{$\sampleIdx = 1 \textrm{ to } \nsamp$}
		\STATE $(\latentS{\sampleIdx}, \paramsS{\sampleIdx}) \gets$ exact sample from $\pmf(\latent, \params \given \obs)$
		\STATE $\weightS{\sampleIdx} \gets 1$
	\ENDFOR
	\FOR{$\dataIdx = \ndist \textrm{ to } 1$}
		\FOR{$\sampleIdx = 1 \textrm{ to } \nsamp$}
			\STATE $(\latentIS{1:\dataIdx}{\sampleIdx}, \paramsS{\sampleIdx}) \gets$ MCMC transition which leaves $\pmf(\latentI{1:\dataIdx}, \params \given \obsI{1:\dataIdx})$ invariant
			\STATE $\weightS{\sampleIdx} \gets \weightS{\sampleIdx} \pmf(\latentIS{\dataIdx}{\sampleIdx} \given \params) \pmf(\obsI{\dataIdx} \given \latentIS{\dataIdx}{\sampleIdx}, \paramsS{\sampleIdx}) / \pmfProposal(\latentI{\dataIdx} \given \obsI{\dataIdx}, \paramsS{\sampleIdx})$
		\ENDFOR
		\IF{resampling criterion met}
			\STATE Resample $(\latentIS{1:\dataIdx}{\sampleIdx}, \paramsS{\sampleIdx})$ proportionally to $1/\weightS{\sampleIdx}$
			\STATE $\sumOfWeights \gets \sum_{\sampleIdx=1}^\nsamp 1 / \weightS{\sampleIdx}$
			\FOR{$\sampleIdx = 1 \textrm{ to } \nsamp$}
				\STATE $\weightS{\sampleIdx} \gets \nsamp / \sumOfWeights$
			\ENDFOR
		\ENDIF
	\ENDFOR
	\RETURN $\pfnEstimate = \frac{\nsamp}{\sum_{\sampleIdx=1}^\nsamp 1/\weightS{\sampleIdx}}$
\end{small}
\end{algorithmic}
\caption{Sequential harmonic mean estimator (SHME)}
\label{alg:shme}
\end{algorithm}

The na\"{\i}ve harmonic mean estimator has been criticized for its instability \citep{neal-worst-monte-carlo}, so it is worth examining whether the same issues are relevant to SHME. One problem with the na\"{\i}ve HME is that it is an instance of simple importance sampling where the target distribution is more spread out than the proposal distribution. Therefore, samples corresponding to the tails of the target distribution can have extremely large importance weights. The same problem is present in SHME (though to a lesser degree) because the weight update steps are based on importance sampling where the proposal distribution is the posterior  $\pmf(\latentI{1:\dataIdx}, \params \given \obsI{1:\dataIdx})$, and the target distribution is the slightly less peaked posterior  $\pmf(\latentI{1:\dataIdx}, \params \given \obsI{1:\dataIdx-1})$ (which conditions on one less data point). Therefore, the individual weight updates, as well as the final output, may have large variance when the algorithm is interpreted as an estimator of $\pmf(\obs)$. 

Yet, when used as an estimator of $\log \pmf(\obs)$, SHME can still yield accurate estimates. To justify this theoretically, suppose we have a model for which $\params$ and $\latentIS{\dataIdx}{\sampleIdx}$ can both be integrated out analytically in the predictive distribution $\pmf(\obsI{\dataIdx} \given \latentIS{1:\dataIdx-1}{\sampleIdx}, \obsI{1:\dataIdx-1})$. (E.g., this is the case for the clustering model we consider in our experiments.) We analyze the gap between the SMC (stochastic lower bound) and SHME (stochastic upper bound) estimates. For simplicity, assume that only a single particle is used in each algorithm, and that the MCMC transition operator yields exact samples in each step. (These correspond to assumptions made by \citet{ais} when analyzing AIS.) In this case, the expected SMC estimate of $\log \pmf(\obs)$ is given by:
\begin{align}
\expect[\log \pmfEstimate_{\rm SMC}(\obs)] &= \sum_{\dataIdx=1}^\ndata \expect_{\pmf(\latentI{1:\dataIdx-1} \given \obsI{1:\dataIdx-1})} \left[ \log \pmf(\obsI{\dataIdx} \given \latentI{1:\dataIdx-1}, \obsI{1:\dataIdx-1}) \right]. \label{eqn:shme-theory-smc}
\end{align}
On the other hand, the expected SHME estimate is given by:
\begin{align}
\expect[\log \pmfEstimate_{\rm SHME}(\obs)] &= \sum_{\dataIdx=1}^\ndata \expect_{\pmf(\latentI{1:\dataIdx-1} \given \obsI{1:\dataIdx})} \left[ \log \pmf(\obsI{\dataIdx} \given \latentI{1:\dataIdx-1}, \obsI{1:\dataIdx-1}) \right]. \label{eqn:shme-theory-shme}
\end{align}
(The difference between these two equations is that the distribution in Eqn.~\ref{eqn:shme-theory-shme} conditions on one additional data point.) Since $\expect[\log \pmfEstimate_{\rm SMC}(\obs)] \leq \log \pmf(\obs) \leq \expect[\log \pmfEstimate_{\rm SHME}(\obs)]$, we can bound the estimation error:
\begin{align}
\left| \expect[\log \pmfEstimate_{\rm SHME}(\obs)] - \log \pmf(\obs) \right| &\leq \expect[\log \pmfEstimate_{\rm SHME}(\obs)] - \expect[\log \pmfEstimate_{\rm SMC}(\obs)] \\
&= \sum_{\dataIdx=1}^\ndata \expect_{\pmf(\latentI{1:\dataIdx-1} \given \obsI{1:\dataIdx})} \left[ \log \pmf(\obsI{\dataIdx} \given \latentI{1:\dataIdx-1}, \obsI{1:\dataIdx-1}) \right] \nonumber \\ &\phantom{=} - \expect_{\pmf(\latentI{1:\dataIdx-1} \given \obsI{1:\dataIdx-1})} \left[ \log \pmf(\obsI{\dataIdx} \given \latentI{1:\dataIdx-1}, \obsI{1:\dataIdx-1}) \right]. \label{eqn:shme-theory-terms}
\end{align}
To understand the terms in this sum, suppose we are predicting the next observation $\obsI{\dataIdx}$ given our past observations. If we ``cheat'' by using $\obsI{\dataIdx}$ to help infer the latent variables for past observations, we would expect this to improve the predictive likelihood. Each of the terms in Eqn.~\ref{eqn:shme-theory-terms} corresponds to the magnitude of this improvement. This is, however, a very indirect way for $\obsI{\dataIdx}$ to influence the model's predictions, so arguably we should expect these terms to be relatively small. If they are indeed small, then SHME will have small error in estimating $\log \pmf(\obs)$.

\subsection{Bidirectional Monte Carlo}

So far, we have discussed techniques for obtaining stochastic lower and upper bounds on the log marginal likelihood for simulated data. The stochastic lower bounds are obtained using standard sampling-based techniques, such as AIS or SMC. The stochastic upper bounds are obtained by running one of these algorithms in reverse, starting from an exact posterior sample. The methods are most useful in combination, since one can sandwich the true log-ML value between the stochastic bounds. Both AIS and SMC are consistent, in the sense that they approach the correct value in the limit of infinite computation \citep{ais,smc}. Therefore, it is possible to run both directions with enough computation that the two stochastic bounds agree. Once the stochastic bounds agree closely (e.g.~to within 1 nat), one has a log-ML estimate which (according to the above analysis based on Markov's inequality) is very unlikely to be off by more than a few nats. Errors of this magnitude are typically inconsequential in the context of log-ML estimation, so one can consider the final log-ML estimate to be ground truth. We refer to this overall approach to computing ground truth log-ML values as bidirectional Monte Carlo (BDMC).

\subsection{Relationship with RAISE}

Our BDMC method is similar in spirit to the reverse AIS estimator \citep{raise}, which has been used to evaluate log-likelihoods of Markov random fields. Like BDMC, RAISE runs AIS both forwards and backwards in order to sandwich the log-likelihoods between two values. The main differences between the methods are as follows:
\begin{enumerate}
\item RAISE is used in the setting of Markov random fields, or undirected graphical models. Algorithms such as AIS compute stochastic lower bounds on the partition function, which correspond to stochastic upper bounds on the log-likelihood. The main difficulty was lower bounding the log-likelihoods.

BDMC is applied to models described as generative processes (and often represented with \emph{directed} graphical models). The challenge is to integrate out parameters and latent variables in order to compute the likelihood of observations. Prior methods often returned stochastic \emph{lower} bounds on the log-likelihood, and the technical novelty concerns stochastic \emph{upper} bounds.

\item RAISE runs the reverse AIS chain starting from the data on which one wishes to evaluate log-likelihoods. BDMC runs the reverse chain starting from an exact posterior sample.

\item RAISE computes stochastic log-likelihood lower bounds for an \emph{approximate} model. If the approximate model describes the data better than the original MRF, RAISE's supposed lower bound may in fact overestimate the log-likelihood. By contrast, BDMC returns stochastic upper and lower bounds on the \emph{original model}, so one can locate the true value between the bounds with high probability.

\end{enumerate}

\subsection{Evaluating posterior inference}
\label{sec:relationship-with-inference}

So far, the discussion has focused on measuring the accuracy of log-ML estimators. In some cases, BDMC can also be used to quantitatively measure the quality of an approximate posterior sampler on simulated data. To do this, we make use of a relationship between posterior inference and marginal likelihood estimation which holds for some sampling-based inference algorithms.

It is well known that the problems of inference and ML estimation are equivalent for variational Bayes (Section~\ref{sec:variational-bayes}): the KL divergence between the approximate and true posteriors equals the gap between the variational lower bound and the true log-ML. A similar relationship holds for some sampling-based log-ML estimators, except that the equality may need to be replaced with an inequality. First, consider the case of simple importance sampling (Section~\ref{sec:likelihood-weighting}). If $\pmfProposal(\latent, \params)$ is the proposal distribution, then the expected log-ML estimate based on a single sample (see Eqn.~\ref{eqn:sis-estimator}) is given by:
\begin{align}
\expect_{\pmfProposal(\latent, \params)} \left[ \log \pmf(\latent, \params, \obs) - \log \pmfProposal(\latent, \params) \right] &= \log \pmf(\obs) + \expect_{\pmfProposal(\latent, \params)} \left[ \log \pmf(\latent, \params \given \obs) - \log \pmfProposal(\latent, \params) \right] \nonumber \\
&= \log \pmf(\obs) - \kldiv(\pmfVariational(\latent, \params) \klBars \pmf(\latent, \params \given \obs)).
\end{align}
Interestingly, this formula is identical to the variational Bayes lower bound when $\pmfProposal$ is used as the approximating distribution. 

We have discussed two sampling-based ML estimators which can be seen as importance sampling on an extended state space: AIS (Section \ref{sec:background-ais}) and SMC with a single particle (Section \ref{sec:relationship-smc-ais}). Let $\auxiliary$ denote all of the variables sampled in one of these algorithms other than $\latent$ and $\params$. (For instance, in AIS, it denotes all of the states other than the final one.) In this case, the above derivation can be modified:
\begin{align}
\expect[\log \pmfEstimate(\obs)] &= \expect_{\pmfProposal(\latent, \params, \auxiliary)} \left[ \log \pmf(\latent, \params, \auxiliary, \obs) - \log \pmfProposal(\latent, \params, \auxiliary) \right] \nonumber \\
&= \log \pmf(\obs) - \kldiv(\pmfProposal(\latent, \params, \auxiliary) \klBars \pmf(\latent, \params, \auxiliary \given \obs)) \nonumber \\
&\leq \log \pmf(\obs) - \kldiv(\pmfVariational(\latent, \params) \klBars \pmf(\latent, \params \given \obs)). \label{eqn:inference-inequality}
\end{align}
This implies that the KL divergence of the approximate posterior samples from the true posterior is bounded by the bias of the log-ML estimator. Eqn.~\ref{eqn:inference-inequality}, in conjunction with BDMC, can be used to demonstrate the accuracy of posterior samples on simulated datasets. In particular, to measure the accuracy of AIS or SMC, one can compute the gap between its log-ML estimate and the stochastic upper bound from BDMC. This gap will be a stochastic upper bound on the KL divergence of the distribution of approximate samples from the true posterior.

\section{Other marginal likelihood estimators}
\label{sec:other-ml-estimators}

In this section, we overview some additional ML estimation algorithms not discussed in Section~\ref{sec:background}.

\subsection{Variational Bayes}
\label{sec:variational-bayes}

All of the methods described above are sampling-based estimators. Variational Bayes \citep{hinton-van-camp,waterhouse-variational-bayes,variational-bayes,variational-bayes-propagation} is an alternative set of techniques based on optimization. In particular, the aim is to approximate the intractable posterior distribution $\pmf(\latent, \params \given \obs)$ with a tractable approximation $\pmfVariational(\latent, \params)$, \emph{i.e.}~one whose structure is simple enough to represent explicitly. Typically, $\latent$ and $\params$ are constrained to be independent, \emph{i.e.}~$\pmfVariational(\latent, \params) = \pmfVariational(\latent) \pmfVariational(\params)$, and the two factors may themselves have additional factorization assumptions. The objective function being maximized is the following:
\begin{align}
\variationalFunc(\pmfVariational) &\triangleq \expect_{\pmfVariational(\latent, \params)}\left[ \log \pmf(\params, \latent, \obs) \right] + \entropy \left[ \pmfVariational(\latent, \params) \right], \label{eqn:variational-bayes-objfn}
\end{align}
where $\entropy$ denotes entropy. This functional is typically optimized using a coordinate ascent procedure, whereby each factor of $\pmfVariational$ is optimized given the other factors. Assuming the factorization given above, the update rules which optimize Eqn \ref{eqn:variational-bayes-objfn} are:
\begin{align}
\pmfVariational(\latent) &\propto \exp \left( \expect_{\pmfVariational(\params)} \left[ \log \pmf(\latent, \params, \obs) \right] \right) \\
\pmfVariational(\params) &\propto \exp \left( \expect_{\pmfVariational(\latent)} \left[ \log \pmf(\latent, \params, \obs) \right] \right)
\end{align}

Variational Bayes is used for both posterior inference and marginal likelihood estimation, and the two tasks are equivalent, according to the following identity:
\begin{align}
\log \variationalFunc(\pmfVariational) &= \log \pmf(\obs) - \kldiv(\pmfVariational(\latent, \params) \klBars \pmf(\latent, \params \given \obs)). \label{eqn:variational-bayes-identity}
\end{align}
\emph{I.e.}, variational Bayes underestimates the true log marginal likelihood, and the gap is determined by the KL divergence from the true posterior.

\subsection{Chib-style estimators}
\label{sec:background-chib-style}

Another estimator which is popular because of its simplicity is Chib's method \citep{chibs-method}. This method is based on the identity
\begin{align}
\pmf(\obs) &= \frac{\pmf(\latentPointEstimate, \paramsPointEstimate, \obs)}{\pmf(\latentPointEstimate, \paramsPointEstimate \given \obs)} \label{eqn:chib-identity}
\end{align}
for any particular values $(\latentPointEstimate, \paramsPointEstimate)$ of the latent variables and parameters. While (\ref{eqn:chib-identity}) holds for any choice of $(\latentPointEstimate, \paramsPointEstimate)$, they are usually taken to be high probability locations, such as the maximum a posteriori (MAP) estimate. The numerator can generally be computed from the model definition. The denominator is based on a Monte Carlo estimate of the conditional probability obtained from posterior samples $(\latentS{1}, \paramsS{1}), \ldots, (\latentS{\nsamp}, \paramsS{\nsamp})$. In particular, let $\trans$ represent an MCMC operator which leaves $\pmf(\latent, \params \given \obs)$ invariant; the basic version of the algorithm assumes a Gibbs sampler. For models where the Gibbs transitions can't be computed exactly, another variant uses Metropolis-Hastings instead \citep{chibs-method-mh}. (The posterior samples may be obtained from a Markov chain using $\trans$, but this is not required.) The denominator is estimated as:
\begin{align}
\pmfEstimate(\latentPointEstimate, \paramsPointEstimate \given \obs) &= \frac{1}{\nsamp} \sum_{\sampleIdx=1}^\nsamp \trans(\latentPointEstimate, \paramsPointEstimate \given \latentS{\sampleIdx}, \paramsS{\sampleIdx}, \obs). \label{eqn:chib-denominator-estimate}
\end{align}

How should the estimator be expected to perform?  Observe that if exact samples are used, (\ref{eqn:chib-denominator-estimate}) is an unbiased estimate of the denominator of (\ref{eqn:chib-identity}). Therefore, following the analysis of Section \ref{sec:background-pfn-principles}, it would tend to underestimate the denominator, and therefore overestimate the true marginal likelihood value. If approximate posterior samples are used, nothing can be said about its relationship with the true value. In this review, we focus on latent variable models, which generally have symmetries corresponding to relabeling of latent components or dimensions. Since transition probabilities between these modes are very small, the estimator could drastically overestimate the marginal likelihood unless the posterior samples happen to include the correct mode. Accounting for the symmetries in the algorithm itself can be tricky, and can cause subtle bugs \citep{neal-letter-chib}.

\citet{chib-style} proposed a variant on Chib's method which yields an unbiased estimate of the marginal likelihood. We will refer to the modified version as the Chib-Murray-Salakhutdinov (CMS) estimator. The difference is that, rather than allowing an arbitrary initialization for the Markov chain over $(\latent, \params)$, they initialize the chain with a sample from $\transReverse(\latent, \params \given \latentPointEstimate, \paramsPointEstimate)$, where
\begin{equation}
\transReverse(\latent^\prime, \params^\prime \given \latent, \params) \triangleq \frac{\trans(\latent, \params \given \latent^\prime, \params^\prime)\, \pmf(\latent^\prime, \params^\prime \given \obs)}{\sum_{\latent^\prime, \params^\prime} \trans(\latent, \params \given \latent^\prime, \params^\prime)\, \pmf(\latent^\prime, \params^\prime \given \obs) }
\end{equation}
is the reverse operator of $\trans$.

\subsection{Nested sampling}
\label{sec:nested-sampling}

Nested sampling \citep[NS;][]{nested-sampling} is a marginal likelihood estimator which, like AIS, samples from a sequence of distributions where the strength of the evidence is gradually amplified. In this section, we denote the state as $\genState = (\params, \latent)$, the prior as $\nsPrior(\genState) = \pmf(\params, \latent)$, and the likelihood as $\nsLikelihood(\genState) = \pmf(\obs \given \params, \latent)$. Central to the method is the notion of the \emph{constrained prior}, which is the prior distribution restricted to a region of high likelihood:
\begin{align}
\nsConstrainedPrior{\nsCutoff}(\genState) &\triangleq
\begin{cases}
\nsPrior(\genState) / \nsVolume(\nsCutoff) & \nsLikelihood(\genState) > \nsCutoff \\
0 & \nsLikelihood(\genState) \leq \nsCutoff
\end{cases} \\
\nsVolume(\nsCutoff) &\triangleq \nsPrior(\{\genState : \nsLikelihood(\genState) > \nsCutoff\}).
\end{align}
The \emph{prior volume} $\nsVolume(\nsCutoff)$ is the fraction of the prior probability mass which lies within the likelihood constraint. For simplicity, we assume the set $\{\genState : \nsLikelihood(\genState) = \nsCutoff\}$ has measure zero for any $\nsCutoff$. Each step of NS attempts to sample from a constrained prior, where the cutoff $\nsCutoff$ is increased (and hence the volume $\nsVolume(\nsCutoff)$ is decreased) at a controlled rate.

We first describe the idealized version of NS, where one assumes an oracle that returns an exact sample from the constrained prior. One begins with a set of $\nsamp$ particles $\{\genStateS{\sampleIdx}\}_{\sampleIdx=1}^\nsamp$ drawn \emph{i.i.d.}~from the prior $\nsPrior$, with $\nsamp \geq 2$. In each step (indexed by $\nsStepIdx$), one chooses the particle with the smallest likelihood; call its index $\smallestSampleIdx \triangleq \arg \min_\sampleIdx  \nsLikelihood(\genStateS{\sampleIdx})$. The likelihood cutoff is updated to $\nsCutoffT{\nsStepIdx} = \nsLikelihood(\genStateS{\smallestSampleIdx})$. The particle $\genStateS{\smallestSampleIdx}$ is then replaced with a sample from the constrained prior $\nsConstrainedPrior{\nsCutoffT{\nsStepIdx}}$. All of the other particles remain untouched. This process is repeated until a stopping criterion (described below) is met.

After step $\nsStepIdx$ of the algorithm, the $\nsamp$ particles are independently distributed according to $\nsConstrainedPrior{\nsCutoffT{\nsStepIdx}}$. The prior volumes $\nsVolume(\nsLikelihood(\genStateS{\sampleIdx}))$ are therefore uniformly distributed on the interval $[0, \nsCutoffT{\nsStepIdx}]$. (By convention, $\nsCutoffT{0} = 0$.) The particle $\genStateS{\smallestSampleIdx}$ (which was chosen to have the minimum likelihood) has a prior volume of approximately $\nsVolume(\nsCutoffT{\nsStepIdx}) \cdot \nsamp / (\nsamp + 1)$. Hence, the prior volume is expected to decrease by roughly a factor of $\nsamp / (\nsamp + 1)$ in each iteration. Since $\nsVolume(\nsCutoffT{0}) = \nsVolume(0) = 1$, this implies
\begin{equation}
\nsVolume(\nsCutoffT{\nsStepIdx}) \approx \left( \frac{\nsamp}{\nsamp+1} \right)^{\nsStepIdx}. \label{eqn:ns-volume-approx}
\end{equation}
Now, observe that for each $\nsStepIdx$, $\nsVolume(\nsCutoffT{\nsStepIdx}) - \nsVolume(\nsCutoffT{\nsStepIdx+1})$ fraction of the prior volume has a likelihood value between $\nsCutoffT{\nsStepIdx}$ and $\nsCutoffT{\nsStepIdx+1}$. Since the marginal likelihood is given by $\pmf(\obs) = \int \nsPrior(\genState) \nsLikelihood(\genState) \,\intD \genState$, we can bound the marginal likelihood above and below:
\begin{equation}
\sum_{\nsStepIdx=0}^\infty (\nsVolume(\nsCutoffT{\nsStepIdx}) - \nsVolume(\nsCutoffT{\nsStepIdx+1}))\, \nsCutoffT{\nsStepIdx}\, \leq\, \pmf(\obs)\, \leq\, \sum_{\nsStepIdx=0}^\infty (\nsVolume(\nsCutoffT{\nsStepIdx}) - \nsVolume(\nsCutoffT{\nsStepIdx+1}))\, \nsCutoffT{\nsStepIdx+1}.
\end{equation}
The true volumes $\nsVolume(\nsCutoffT{\nsStepIdx})$ are unknown, but one can achieve a good approximation by plugging in Eqn~\ref{eqn:ns-volume-approx}. As a stopping criterion, one typically stops when the next term in the summation increases the total by less than a pre-specified ratio (such as $1 + 10^{-10}$).

The idealized algorithm requires the ability to sample exactly from the constrained prior. In practice, this is typically intractable or inefficient. Instead, one tries to approximately sample from the constrained prior using the following procedure: first replace $\genStateS{\smallestSampleIdx}$ with one of the other particles, chosen uniformly at random. Apply one or more steps of MCMC, where the transition operator has the constrained prior as its stationary distribution. If the transition operator mixes fast enough, then this should be a reasonable approximate sample from the constrained prior. 

Like AIS, NS has the interpretation of moving through a series of distributions where the evidence gradually becomes stronger. One of the arguments made for NS, in contrast with AIS, is that its schedule for moving through its space of distributions is potentially more stable \citep{nested-sampling}. In particular, AIS has been observed to suffer from phase transitions: around some critical temperature, the distribution may suddenly change from being very broad to very peaked. If the particles do not quickly find their way into the peak, the results might be inaccurate. In NS, the prior volume decreases at a controlled rate, so this sort of phase transition is impossible. This effect was shown to improve the stability of partition function estimation for Potts models \citep{nested-sampling-potts}.

On the flip side, NS does not have quite as strong a theoretical guarantee as AIS. AIS is guaranteed to be an unbiased estimator of the ML, even when MCMC operators (rather than exact samples) are used in each step. By contrast, the theoretical analysis of the variance of NS \citep{nested-sampling} assumes exact samples in each step. Indeed, in our own experiments, we found that while NS tended to underestimate the ML on average (similarly to AIS), it overestimated the true value too often for it to be a stochastic lower bound in the sense of Section~\ref{sec:background-pfn-principles}.

\section{Experiments}
\label{sec:experiments}

In this section, we use our proposed bidirectional Monte Carlo technique to evaluate a wide variety of ML estimators on several latent variable models. In particular, we consider the following models:
\begin{itemize}
\item {\bf Clustering.} Roughly speaking, this model is a Bayesian analogue of K-means. Each data point is assumed to be drawn from one of $\ncomp$ mixture components. Each mixture component is associated with a spherical Gaussian distribution whose variance is fixed but whose mean is unknown. Mathematically,
\begin{align*}
z_i &\sim {\rm Multinomial}({\boldsymbol \pi}) \\
\theta_{kj} &\sim \normal(0, \sigma^2_\theta) \\
y_{ij} &\sim \normal(\theta_{z_i,j}, \sigma^2_n).
\end{align*}
The mixture probabilities ${\boldsymbol \pi}$, the between-cluster variance $\sigma^2_\theta$, and the within-cluster variance $\sigma^2_n$ are all fixed. In the matrix decomposition grammar of \citet{uai2012}, this model would be written as $\mComp \gComp + \gComp$.

\item {\bf Low-rank approximation.} In this model, we approximate an $\ndata \times \ndim$ matrix $\obsMat$ as a low rank matrix plus Gaussian noise. In particular, we approximate it with the product ${\bf U}{\bf V}$, where ${\bf U}$ and ${\bf V}$ are matrices of size $\ndata \times \ncomp$ and $\ncomp \times \ndim$, respectively. We assume $\ncomp < \min(\ndata, \ndim)$, so the product has rank $\ncomp$. We assume a spherical Gaussian observation model, as well as spherical Gaussian priors on the components ${\bf U}$ and ${\bf V}$. More precisely,
\begin{align*}
u_{ik} &\sim \normal(0, \sigma^2_u) \\
v_{kj} &\sim \normal(0, \sigma^2_v) \\
y_{ij} &\sim \normal({\bf u}_i^T {\bf v}_j, \sigma^2_n).
\end{align*}
The prior variances $\sigma^2_u$ and $\sigma^2_v$ and noise variance $\sigma^2_n$ are all fixed. This model is roughly equivalent to probabilistic matrix factorization \citep{pmf}, except that in our experiments, $\obsMat$ is fully observed. While the model is symmetric with respect to rows and columns, we follow the convention where rows of $\obsMat$ represent data points. In this setup, we can think of the component ${\bf V}$ as the parameters of the model and ${\bf U}$ as the latent variables. In the grammar of \citet{uai2012}, this model would be written as $\gComp \gComp + \gComp$.

\item {\bf Binary attributes.} This model assumes each data point can be described in terms of $\ncomp$ binary-valued attributes. In particular, we approximate the observation matrix $\obsMat$ with the product ${\bf Z}{\bf A}$, where ${\bf Z}$ is an $\ndata \times \ncomp$ binary-valued matrix and ${\bf A}$ is a $\ncomp \times \ndim$ real-valued matrix. Each row of $\obsMat$ corresponds to a single data point. Each row of ${\bf Z}$ can be thought of as a latent vector explaining a data point, and each row of ${\bf A}$ can be thought of as a set of parameters describing the effect one of the binary attributes has on the observations. Mathematically, this model is defined as
\begin{align*}
z_{ik} &\sim {\rm Bernoulli}(\pi_k) \\
a_{kj} &\sim \normal(0, \sigma^2_a) \\
y_{ij} &\sim \normal(\sum_k z_{ik} a_{kj}, \sigma^2_n).
\end{align*}
The attribute probabilities $\{\pi_k\}$, feature variance $\sigma^2_a$, and noise variance $\sigma^2_n$ are all fixed. This model is related to the Indian buffet process linear-Gaussian model \citep{ibp}, with the important difference that both the number of attributes and the probability of each one are fixed. In the grammar of \citet{uai2012}, this model would be written as $\bComp \gComp + \gComp$.
\end{itemize}
We note that all of the models described above have hyperparameters, such as mixture probabilities or noise variance. In a practical setting, these hyperparameters are typically unknown. One would typically include them as part of the model, using appropriate priors, and attempt to infer them jointly with the model parameters and latent variables. Unfortunately, this does not work well in our setting, due to our need to simulate data from the model. One ordinarily assigns weak priors to the hyperparameters, but sampling from such priors usually results in pathological datasets where the structure is either too weak to detect or so strong that it is trivial to find the correct explanation. To avoid these pathologies, we assigned fixed values to the hyperparameters, and we chose these values such that the posterior distribution captures most of the structure, but is not concentrated on a single explanation.

We evaluated the following ML estimators on all three of these models:
\begin{itemize}
\item the Bayesian information criterion (BIC)
\item likelihood weighting (Section~\ref{sec:likelihood-weighting})
\item the harmonic mean estimator (HME) (Section~\ref{sec:harmonic-mean-estimator}), using a Markov chain starting from the exact sample
\item annealed importance sampling (AIS) (Section \ref{sec:background-ais})
\item sequential Monte Carlo (SMC), using a single particle (Section~\ref{sec:particle-filter})
\item variational Bayes (Section~\ref{sec:variational-bayes}). We report results both with and without the symmetry coorection, where the ML lower bound is multiplied by the number of equivalent relabelings ($\ncomp!$ for all models we consider).
\item the Chib-Murray-Salakhutdinov (CMS) estimator (Section~\ref{sec:background-chib-style})
\item nested sampling (Section~\ref{sec:nested-sampling})
\end{itemize}

\subsection{Implementation}

In order to make the running times of different algorithms directly comparable, the implementations share the same MCMC transition operators wherever possible. The only exceptions are variational Bayes and nested sampling, whose update rules are not shared with the other algorithms.

All of the estimators except for BIC, likelihood weighting, and variational Bayes require an MCMC transition operator which preserves the posterior distribution. In addition, some of the algorithms require implementing some additional computations:
\begin{itemize}
\item AIS requires MCMC operators for each of the intermediate distributions. 
\item SMC requires the ability to compute or approximate the likelihood of a data point under the predictive distribution.
\item The CMS estimator requires implementing the reverse transition operators. It also requires computing the transition probabilities between any pair of states. The latter imposes a nontrivial constraint on the choice of MCMC operators, in that it disallows operators which compute auxiliary variables.
\item Nested sampling requires an MCMC operator whose stationary distribution is the constrained prior.
\item Unlike the other algorithms, variational Bayes is based on optimization, rather than sampling. In our implementation, the updates all involved optimizing one of the component distributions given the others.
\end{itemize}

For all three models, the MCMC transition operator was a form of Gibbs sampling. Here are some more model-specific details:
\begin{itemize}
\item {\bf Clustering.} The cluster centers were collapsed out wherever possible in all computations. The predictive likelihood can be computed exactly given the cluster assignments and variance parameters, with the cluster centers collapsed out.
\item {\bf Low rank.} Each of the two factors ${\bf U}$ and ${\bf V}$ was resampled as a block. For computing predictive likelihood, ${\bf V}$ was sampled from the posterior, and the ${\bf U}$ was marginalized out analytically.
\item {\bf Binary attributes.} The feature matrix ${\bf A}$ was collapsed out wherever possible, and the tricks of \citet{accelerated-ibp} were used to efficiently update the posterior distribution over ${\bf A}$. 
\end{itemize}

ML estimators are notoriously difficult to implement correctly, as small bugs can sometimes lead to large errors in the outputs without any obvious indication that something is amiss. (Indeed, checking correctness of MCMC samplers and ML estimators is one potential application of this work.) Appendix~\ref{app:testing} discusses in detail our approach to testing the correctness of the implementations of our algorithms.

In general, we face a tradeoff between performance and difficulty of implementation. Therefore, it is worth discussing the relative difficulty of implementing different estimators. In general, BIC, likelihood weighting, and the harmonic mean estimator required almost no work to implement beyond the MCMC sampler. Of the sampling based estimators, AIS and nested sampling required the most work to implement, because they each required implementing a full set of MCMC transition operators specific to those algorithms.\footnote{AIS is most often used in the undirected setting, where the transition operators for the model itself are easily converted to transition operators for the intermediate distributions. In the directed setting, however, raising the likelihood to a power can destroy the directed structure, and therefore implementing collapsed samplers can be more involved.} SMC and the CMS estimator were in between: they required only a handful of additional functions beyond the basic MCMC operators. 

Compared with the sampling methods, variational Bayes typically required somewhat more math to derive the update rules. However, it was considerably simpler to test (see Appendix~\ref{app:testing}). For the low rank and clustering models, implementing variational Bayes required a comparable amount of effort to implementing the MCMC transitions. For the binary attribute model, variational Bayes was considerably easier to implement than the efficient collapsed sampler.

\subsection{Algorithm parameters}
\label{sec:eval-params}

Each of the ML estimators provides one or more knobs which control the tradeoff between accuracy and computation time. In order to investigate the accuracy as a function of running time, we varied one knob for each algorithm and set the rest to reasonable defaults. The following parameters were varied for each algorithm:
\begin{itemize}
\item {\bf Likelihood weighting} and {\bf harmonic mean:} The independent variable was the number of proposals.

\item {\bf Annealed importance sampling:} The annealing path consisted of geometric averages of the initial and target distributions. Because AIS is sometimes unstable near the endpoints of a linear path, we used the following sigmoidal schedule which allocates more intermediate distributions near the endpoints:
\begin{align*}
\tilde{\pathParam}_\distIdx &= \sigma \left( \delta \left( \frac{2 \distIdx}{\ndist} - 1 \right) \right) \\
\pathParam_\distIdx &= \frac{\tilde{\pathParam}_\distIdx - \tilde{\pathParam}_1}{\tilde{\pathParam}_\ndist - \tilde{\pathParam}_1},
\end{align*}
where $\sigma$ denotes the logistic sigmoid function and $\delta$ is a free parameter. (We used $\delta = 4$.) In our experiments, the independent variable was $\ndist$, the number of intermediate distributions.

\item {\bf Sequential Monte Carlo:} We used only a single particle in all experiments, and the independent variable was the number of MCMC transitions per data point.

\item {\bf Chib-Murray-Salakhutdinov:} We used a single sample $(\params^*, \latent^*)$ and varied the number of MCMC transitions starting from that sample.

\item {\bf Variational Bayes:} The independent variable was the number of random restarts in the optimization procedure. Specifically, in each attempt, optimization was continued until the objective function improved by less than 0.01 nats over 50 iterations, at which point another random restart was done. The highest value obtained over all random restarts was reported.

\item {\bf Nested sampling:} The algorithm has three parameters: the number of steps, the number of particles, and the number of MCMC transitions per step. The number of steps was chosen automatically by stopping when the (multiplicative) likelihood updates dropped below $1 + e^{-10}$. We found that using only 2 particles (the smallest number for which the algorithm is defined) consistently gave the most accurate results for modest computation time. Therefore, the independent variable was the number of MCMC transitions per step.
\end{itemize}

When applying algorithms such as AIS or SMC, it is common to average the estimates over multiple samples, rather than using a single sample. For this set of experiments, we ran 25 independent trials of each estimator. We report two sets of results: the average estimates using only a single sample, and the estimates which combine all of the samples.\footnote{For algorithms which are unbiased estimators of the marginal likelihood (AIS, SMC, CMS, and likelihood weighting), the arithmetic mean of the individual estimates was taken. For algorithms which are unbiased estimators of the reciprocal (harmonic mean, reverse AIS, SHME), the harmonic mean was used. For variational inference, the max over all trials was used. For nested sampling, the average of the log-ML estimates was used.} As discussed in Section~\ref{sec:eval-ml-results}, there was little qualitative difference between the two conditions.

\subsection{How much accuracy is required?}
\label{sec:eval-accuracy}

What level of accuracy do we require from an ML estimator?  At the very least, we would like the errors in the estimates to be small enough to detect ``substantial'' log-ML differences between alternative models. \citet{kass-raftery-ml-review} offered the following table to summarize significance levels of ML ratios:

\vspace{1em}
\begin{small}
\begin{tabular}{ccl}
$\log_{10} \pmf_1(\obs) - \log_{10} \pmf_2(\obs)$ & $\pmf_1(\obs) / \pmf_2(\obs)$ & Strength of evidence against $\pmf_2$ \\
\rule{0pt}{1.5em}
0 to 1/2 & 1 to 3.2 & Not worth more than a bare mention \\
1/2 to 1 & 3.2 to 10 & Substantial \\
1 to 2 & 10 to 100 & Strong \\
$>2$ & $>100$ & Decisive
\end{tabular}
\end{small}
\vspace{1em}

\noindent This table serves as a reference point if one believes one of the models is precisely correct. However, in most cases, all models under consideration are merely simplifications of reality. 

Concretely, suppose we have a dataset consisting of $\ndata = 1000$ data points, and we are considering two models, $\model_1$ and $\model_2$. If $\model_1$ achieves an average predictive likelihood score which is 0.1 nats per data point higher than that of $\model_2$, this translates into a log-ML difference of 100 nats. Interpreted as a log-odds ratio, this would be considered overwhelming evidence. However, the difference in predictive likelihood is rather small, and in practice may be outweighed by other factors such as computation time and interpretability. Roughly speaking, 1 nat is considered a large difference in predictive likelihood, while 0.1 nats is considered small. Therefore, we may stipulate that an ML estimation error of $0.1 \ndata$ nats is acceptable, while one of $\ndata$ nats is not.

Alternatively, there is an empirical yardstick we can use, namely comparing the ML scores of different models fit to the same datasets. Table~\ref{fig:eval-wrong-model} shows the ML estimates for all three models under consideration, on all three of the simulated datasets.\footnote{Unlike in the rest of our experiments, it did not make sense to freeze the model hyperparameters for the cross-model ML evaluation, as there is no ``correct'' choice of hyperparameters when the model is wrong. Therefore, for the results in this table, we included the hyperparameters in the model, and these were integrated out along with the parameters and latent variables. We used standard priors for hyperparameters: inverse gamma distributions for variances, Dirichlet distributions for mixture probabilities, and beta distributions for Bernoulli probability parameters. AIS was generally able to infer the correct hyperparameter values in this experiment.} The estimates were obtained from AIS with 30,000 intermediate distributions.\footnote{The entries in this table are guaranteed only to be stochastic lower bounds. However, AIS with 30,000 intermediate distributions yielded accurate estimates in all comparisons against ground truth (see Section~\ref{sec:eval-ml-results}).} These numbers suggest that, for a dataset with 50 data points and 25 dimensions, the ML estimators need to be accurate to within tens of nats to distinguish different Level 1 factorization models.

\begin{table}
\begin{center}
\begin{tabular}{r@{\hspace{2em}}cc@{\hspace{2em}}cc@{\hspace{2em}}cc}
& \multicolumn{2}{c}{\bf Clustering} & \multicolumn{2}{c}{\bf Low rank} & \multicolumn{2}{c}{\bf Binary} \\
\rule{0pt}{1.3em}
{\bf Clustering} & -2377.5 & & -2390.6 & (13.1) & -2383.2 & (5.7) \\
{\bf Low rank} & -2214.2 & (69.1) & -2145.1 & & -2171.4 & (26.3) \\
{\bf Binary} & -2268.6 & (49.2) & -2241.7 & (22.3) & -2219.4 &
\end{tabular}
\end{center}
\caption{Marginal likelihood scores for all three models evaluated on simulated data drawn from all three models. {\bf Rows:} the model used to generate the simulated data. {\bf Columns:} the model fit to the data. Each entry gives the marginal likelihood score estimated using AIS, and (in parentheses) the log-ML difference from the correct model.}
\label{fig:eval-wrong-model}
\end{table}

\subsection{Results}
\label{sec:eval-ml-results}

All of the ML estimation algorithms were run on all three of the models. A simulated dataset was generated for each model with 50 data points and 25 input dimensions. There were 10 latent components for the clustering and binary models and 5 for the low rank model. In all cases, the ``ground truth'' estimate was obtained by averaging the log-ML estimates of the forward and reverse AIS chains with the largest number of intermediate distributions. In all cases, the two estimates agreed to within 1 nat. Therefore, by the analysis of Section~\ref{sec:background-pfn-principles}, the ground truth value is accurate to within a few nats with high probability. 

As mentioned in Section~\ref{sec:eval-params}, each algorithm was run independently 25 times, and the results are reported both for the individual trials and for the combined estimates using all 25 trials. We plot the average log-ML estimates as a function of running time in order to visualize the bias of each estimator. In addition, we plot the mean squared error (MSE) values as a function of running time. We do not report MSE values for the AIS runs with the largest number of intermediate distributions because the estimates were used to compute the ground truth value.

Section~\ref{sec:eval-accuracy} argued, from various perspectives, that the log-ML estimates need to be accurate on the order of 10 nats to distinguish different model classes. Therefore, for all models, we report which algorithms achieved root mean squared error (RMSE) of less than 10 nats, and how much time they required to do so.

{\bf Clustering.} The results for the clustering model are shown in Figures~\ref{fig:clustering-estimates} and \ref{fig:clustering-mse}. Figure~\ref{fig:clustering-estimates} shows the log-ML estimates for all estimators, while Figure~\ref{fig:clustering-mse} shows the RMSE of the log-ML estimates compared to the ground truth. Of the algorithms which do not require an exact posterior sample, only three achieved the desired accuracy: AIS, SMC, and nested sampling (NS). SMC gave accurate results the fastest, achieving an RMSE of 4.6 nats in only 9.7 seconds. By comparison, AIS took 37.8 seconds for an RMSE of 7.0 nats, and NS took 51.2 seconds for an RMSE of 5.7 nats.

\newcommand{\evalFigW}{0.45 \textwidth}
\newcommand{\evalFigSp}{2em}
\newcommand{\evalFigWSmall}{0.33 \textwidth}
\newcommand{\evalFigSpSmall}{-0.01 \textwidth}

\begin{figure}
\begin{center}
\includegraphics[width=\evalFigW]{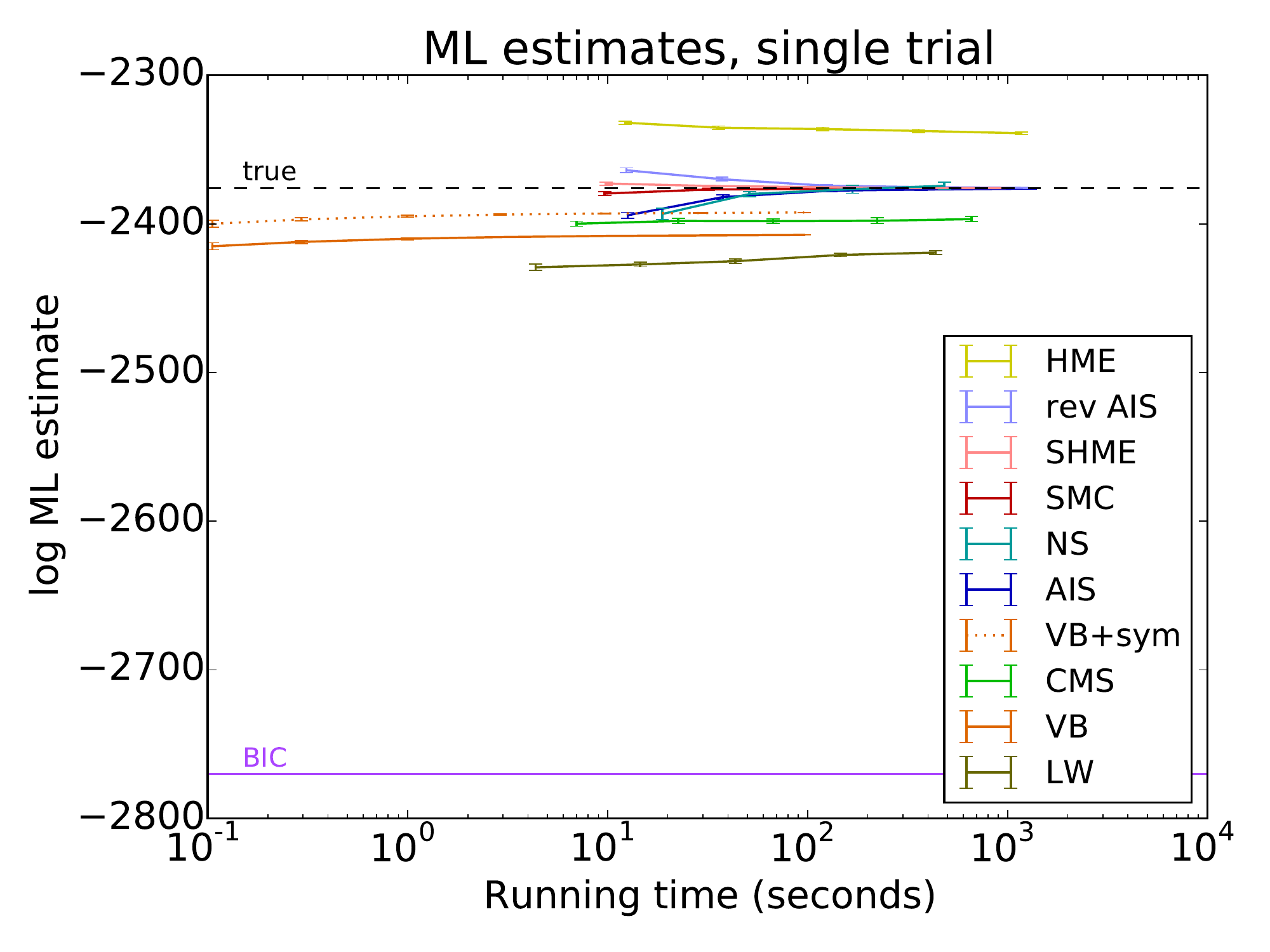}
\hspace{\evalFigSp}
\includegraphics[width=\evalFigW]{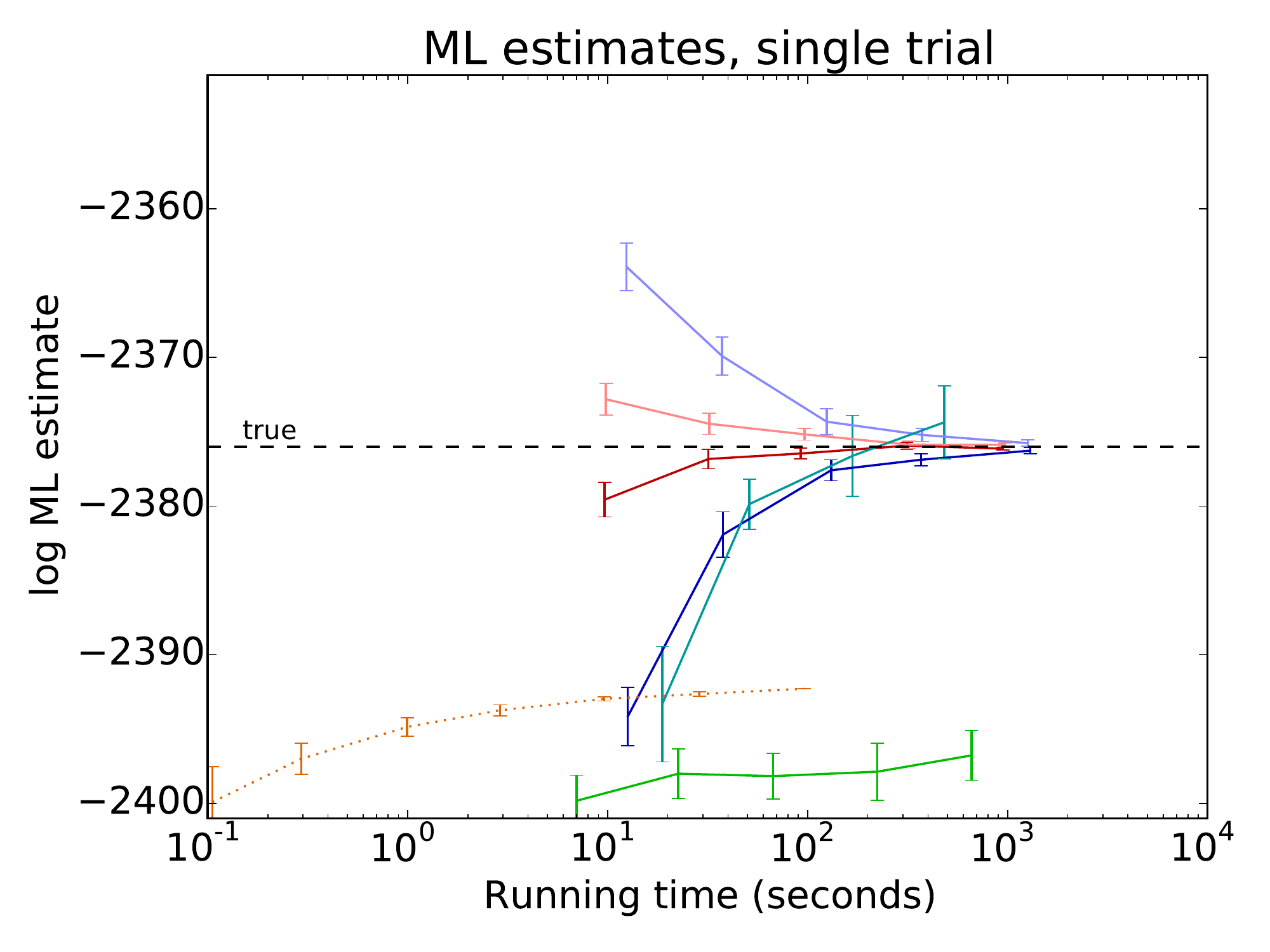} \\
\includegraphics[width=\evalFigW]{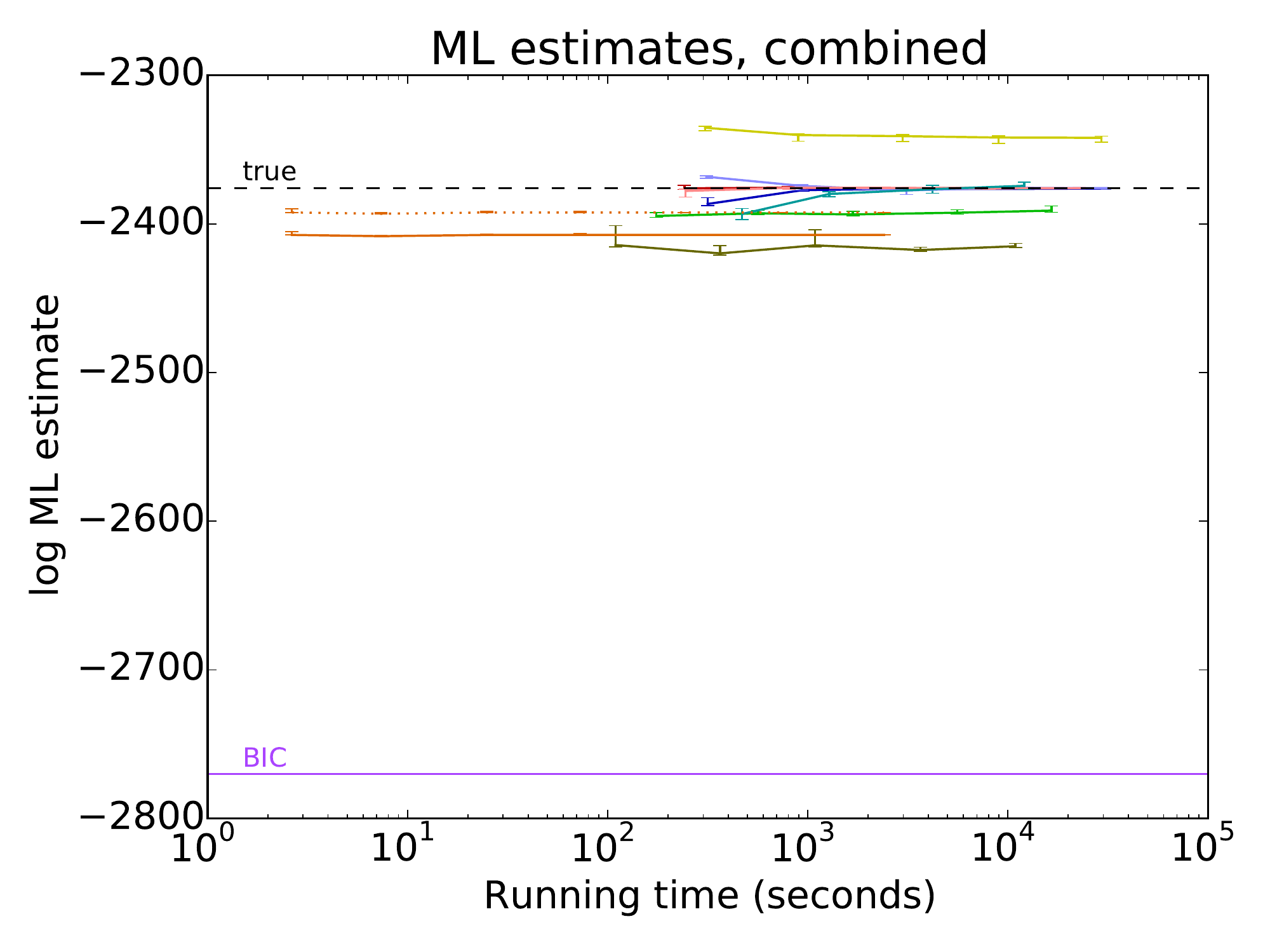}
\hspace{\evalFigSp}
\includegraphics[width=\evalFigW]{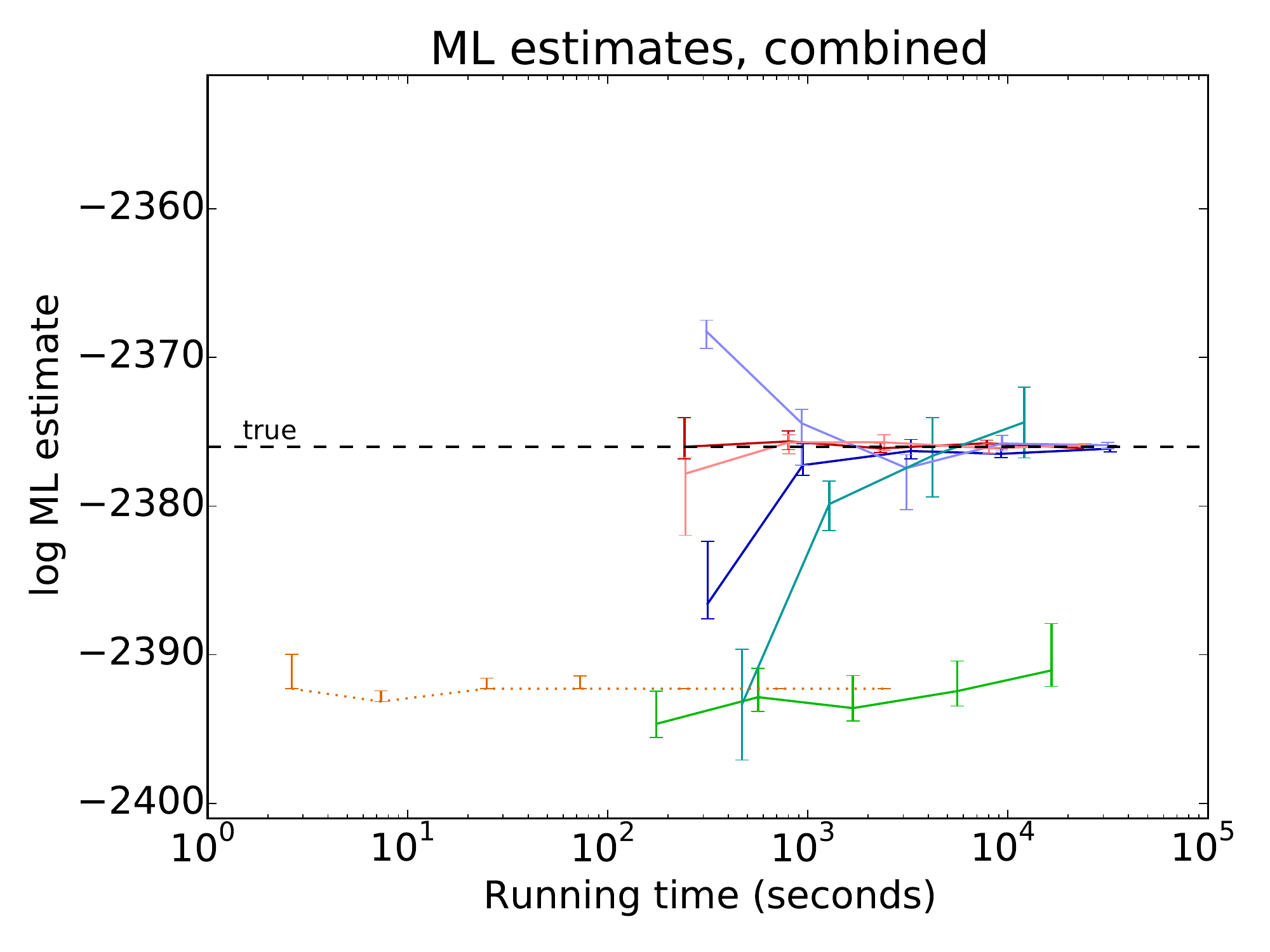}
\end{center}
\caption{Comparison of marginal likelihood estimators on the clustering model. {\bf Top:} average log-ML estimates for each of the 25 individual trials. (The right-hand figure is zoomed in.) {\bf Bottom:} average log-ML estimates combined between the 25 trials. (The right-hand figure is zoomed in.) Note that there is little qualitative difference from the individual trials. {\bf HME} = harmonic mean estimator. {\bf rev AIS} = reverse AIS. {\bf SHME} = sequential harmonic mean estimator. {\bf SMC} = sequential Monte Carlo. {\bf NS} = nested sampling. {\bf AIS} = annealed importance sampling. {\bf VB+sym} = variational Bayes with symmetry correction. {\bf CMS} = Chib-Murray-Salakhutdinov estimator. {\bf VB} = variational Bayes. {\bf LW} = likelihood weighting. Confidence intervals are given for the \emph{expected log-ML estimate} for a given estimator. (They are not confidence intervals for the log-ML itself, so it is not problematic that they generally do not cover the true log-ML.)}
\label{fig:clustering-estimates}
\end{figure}

\begin{figure}
\begin{center}
\includegraphics[width=\evalFigW]{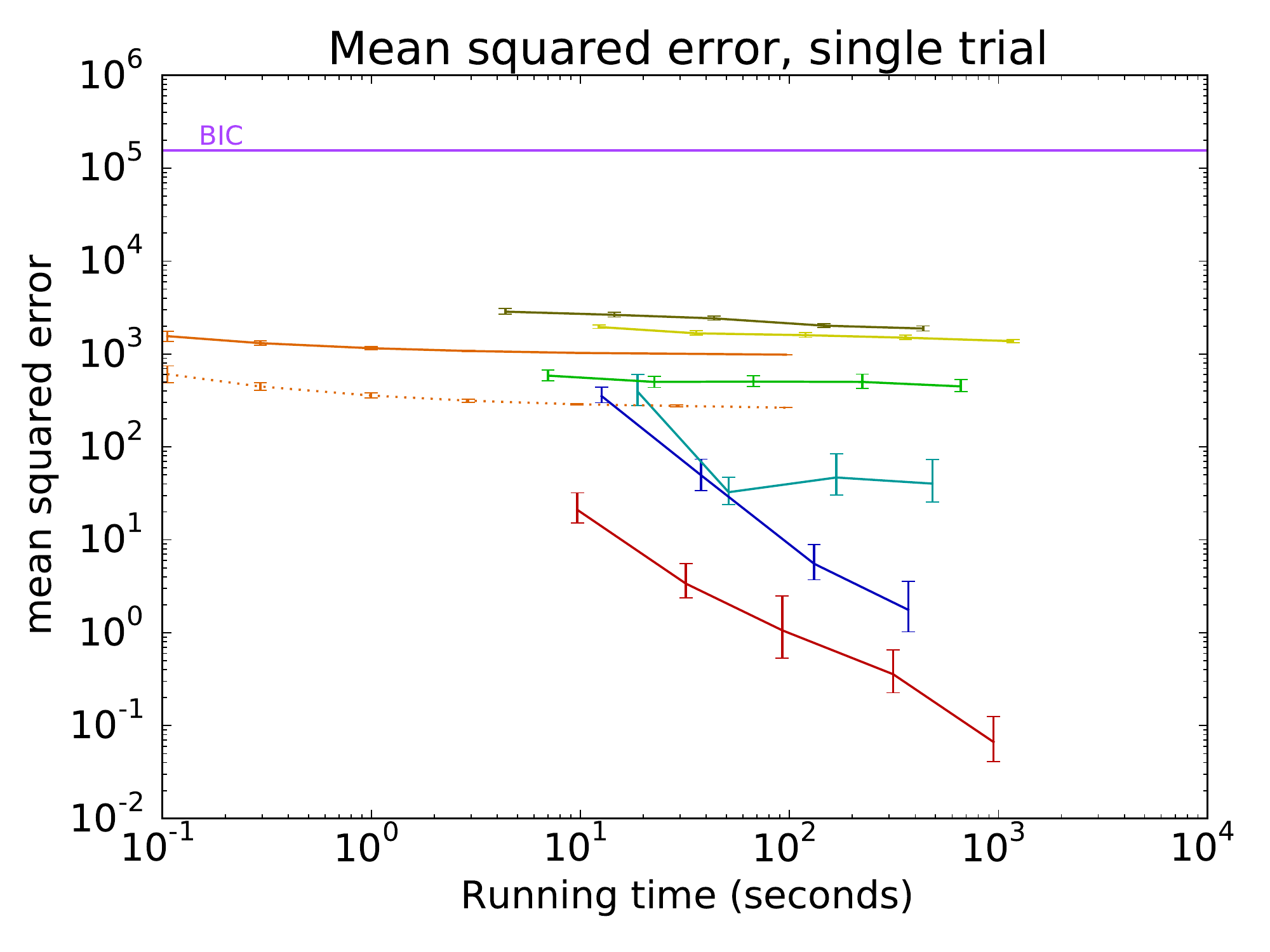}
\hspace{\evalFigSp}
\includegraphics[width=\evalFigW]{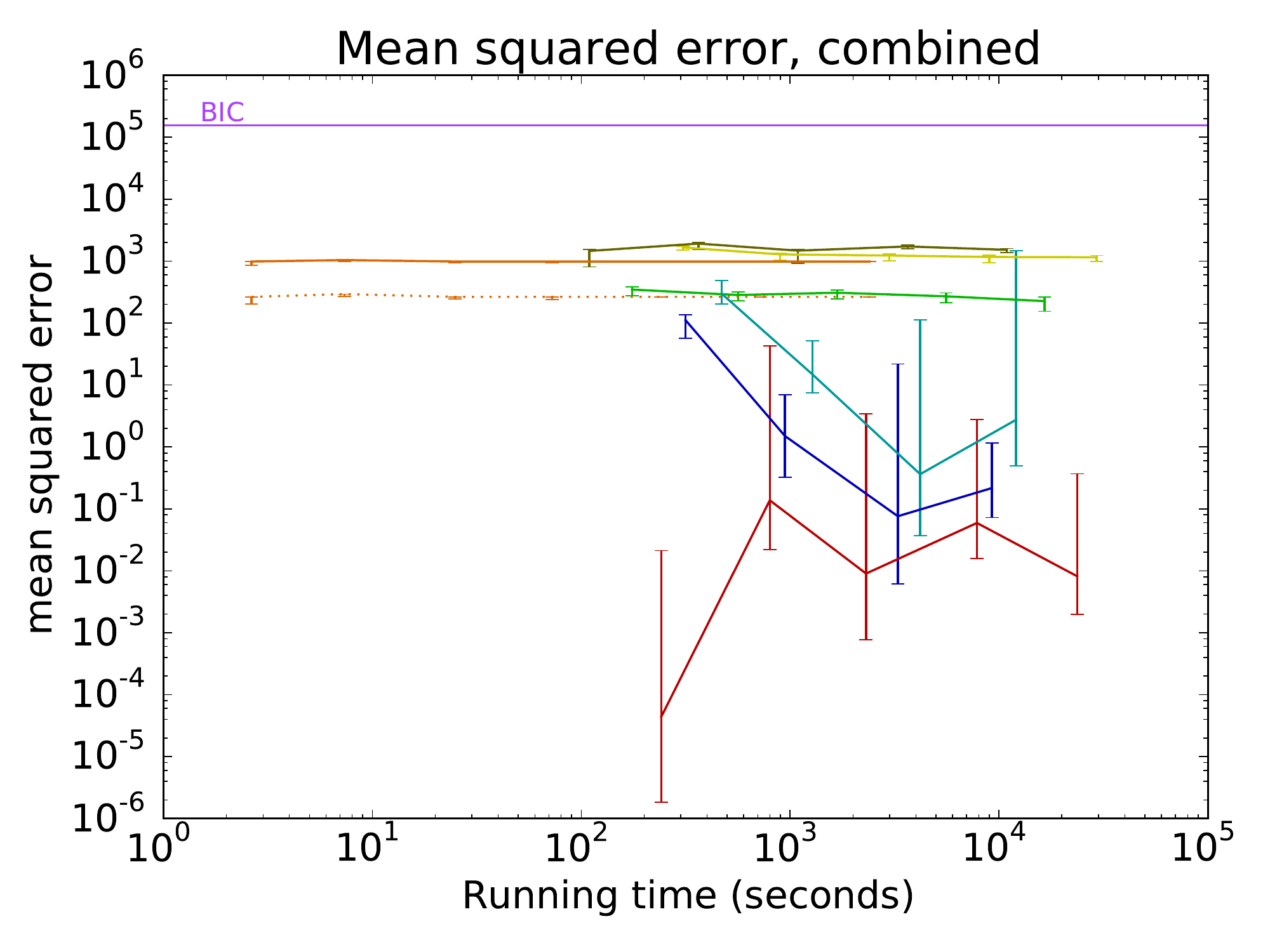}
\end{center}
\caption{Mean squared error relative to ground truth for individual trials (left) and combined estimates (right) for the clustering model. See Figure~\ref{fig:clustering-estimates} for the legend.}
\label{fig:clustering-mse}
\end{figure}

We are not aware of any mathematical results concerning whether NS is an upper or lower bound on the log-ML. Our results suggest that it tends to underestimate the log-ML, similarly to the other algorithms. However, it significantly overestimated the log-ML on many individual runs, suggesting that it is not truly a stochastic lower bound.

The most naive ML estimator, likelihood weighting (LW), vastly underestimated the true value. Its mirror image, the harmonic mean estimator (HME), vastly overestimated it. The Bayesian information criterion (BIC) gave by far the least accurate estimate, with MSE dwarfing even that of LW. This is remarkable, since the BIC requires fitting the model, while LW is simply a form of random guessing. This suggests that the BIC should be treated cautiously on small datasets, despite its asymptotic guarantees. The CMS estimator was more accurate than LW and HME, but still far from the true value. The MSE values for LW, HME, and CMS were nearly constant over at least 2 orders of magnitude in running time, suggesting that these methods cannot be made more accurate simply by running them longer.

A single run of the variational Bayes optimization took only 0.1 seconds, after which it returned a log-ML lower bound which was better than LW achieved after many samples. However, even after many random restarts, the best lower bound it achieved was still 15 nats below the true value, even with the symmetry correction. According to our earlier analysis, this suggests that it would not be accurate enough to distinguish different model classes. In order to determine if the gap was due to local optima, we ran an additional experiment where we gave VB a ``hint'' by initializing it to a point estimate on the sample which generated the data. In this experiment (as well as for the low rank and binary attribute models), VB with random initializations was able to find the same optimum, or a slightly better one, compared with the condition where it was given the hint. This suggests that the gap is due to an inherent limit in the approximation rather than to local optima.

For parameter settings where individual samples of AIS and SMC gave results accurate to within 10 nats, combining the 25 samples gave quantitatively more accurate estimates. However, for all of the other algorithms and parameter settings, combining 25 trials made little difference to the overall accuracy, suggesting that an inaccurate estimator cannot be made into an accurate one simply by using more samples. (The same effect was observed for the low rank and binary attribute models.) Roughly speaking, if one has a fixed computational budget, it is better to compute a handful of accurate estimates rather than a large number of sloppy ones. (Note that this is not true for all partition function estimation problems; for instance, in some of the experiments of \citet{moment-averaging}, high-accuracy results were often obtained by averaging over many AIS runs, even though a large fraction of individual runs gave inaccurate estimates.)

{\bf Low rank.} The results on the low rank factorization overwhelmingly favor AIS: its accuracy after only 1.6 seconds (RMSE = 8.6) matched or surpassed all other algorithms with up to 20 minutes of running time. In fact, AIS was the only algorithm to achieve an RMSE of less than 10.

\begin{figure}
\includegraphics[width=\evalFigWSmall]{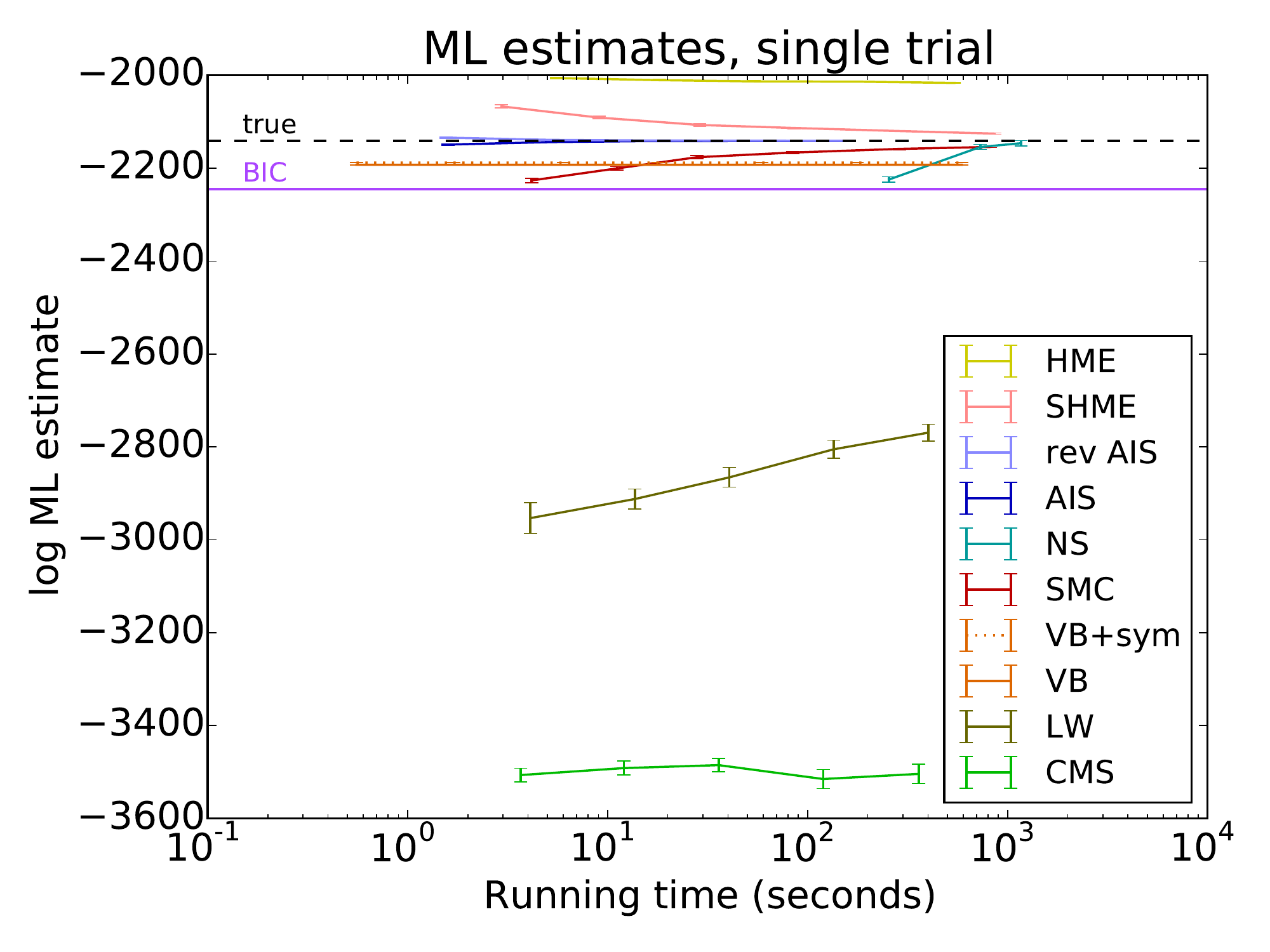}
\hspace{\evalFigSpSmall}
\includegraphics[width=\evalFigWSmall]{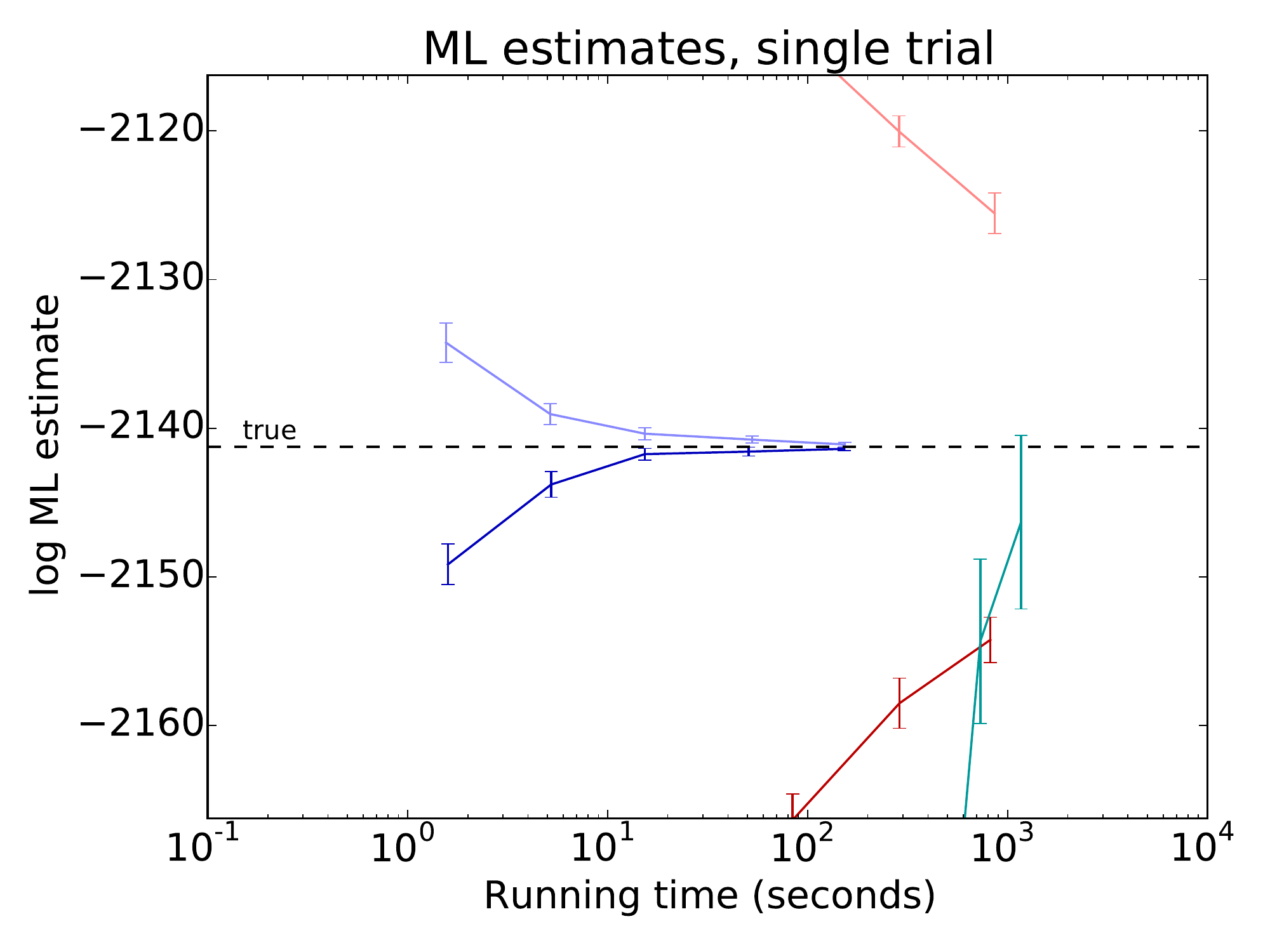}
\hspace{\evalFigSpSmall}
\includegraphics[width=\evalFigWSmall]{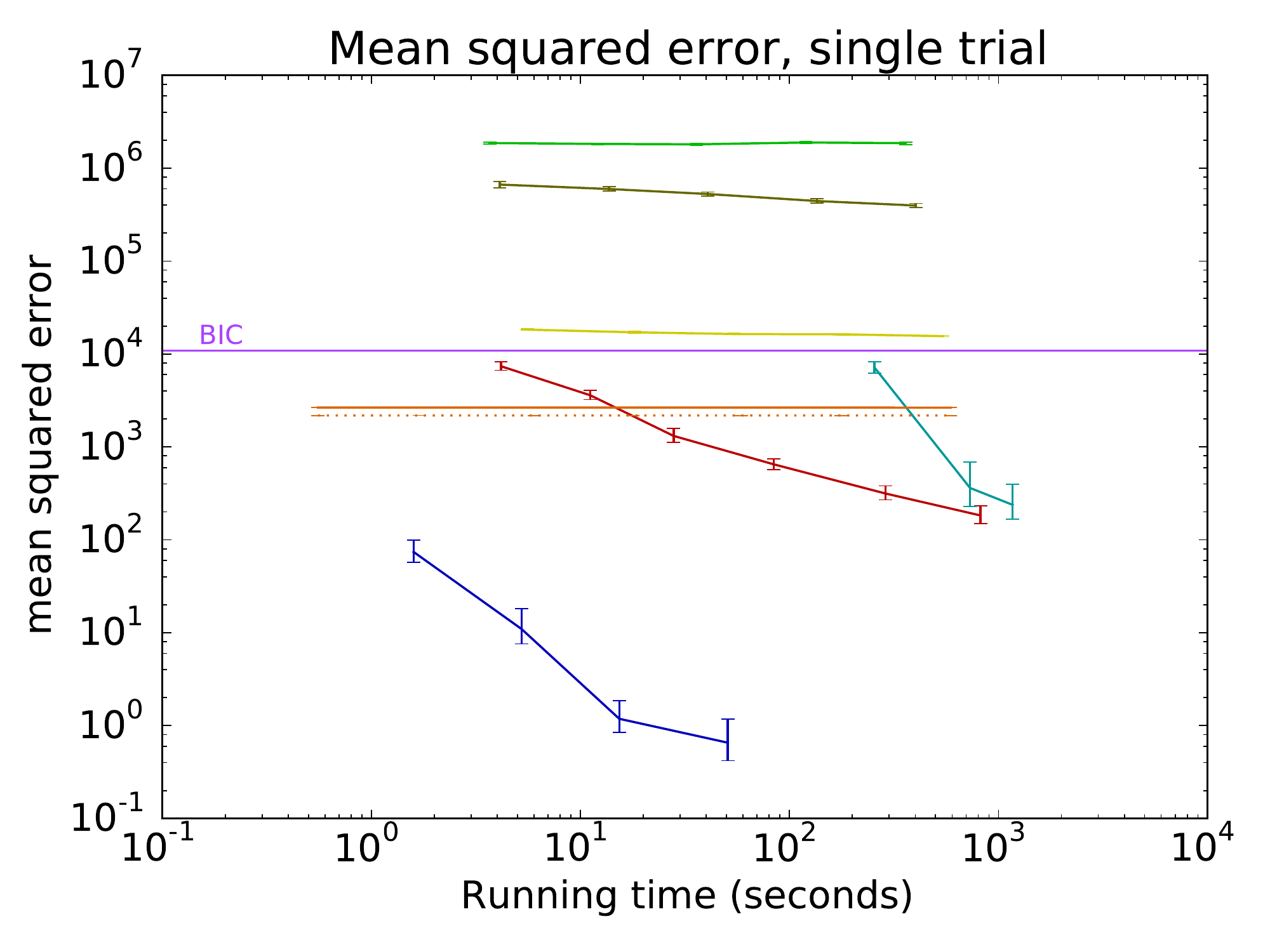}
\caption{Comparison of marginal likelihood estimators on the low rank model. {\bf Left:} average log-ML estimates across the 25 trials. {\bf Middle:} same as left, but zoomed in. {\bf Right:} MSE of individual samples. See Figure~\ref{fig:clustering-estimates} for the abbreviation key.}
\label{fig:low-rank-comparison}
\end{figure}

One reason that NS did not perform as well on this model as it did on the clustering model is that it took more steps to reach the region with high posterior mass. \emph{E.g.}, with 5 MCMC transitions per step, it required 904 steps, as compared with 208 for clustering and 404 for binary. Another reason is that the MCMC implementation could not take advantage of the same structure which allowed block Gibbs sampling for the remaining algorithms; instead, one variable was resampled at a time from its conditional distribution. (For the clustering and binary models, the NS transition operators were similar to the ones used by the other algorithms.)

The CMS estimator underestimated the true value by over 1000 nats. The reason is that $\pmf(\leftFactorPointEstimate, \rightFactorPointEstimate \given \obsMat)$ was estimated using an MCMC chain starting close to the point estimate $\pmf(\leftFactorPointEstimate, \rightFactorPointEstimate)$. The model has a large space of symmetries which the Markov chain explored slowly, because $\leftFactor$ and $\rightFactor$ were tightly coupled. Therefore, the first few samples dramatically overestimated the probability of transitioning to $(\leftFactorPointEstimate, \rightFactorPointEstimate)$, and it was impossible for later samples to cancel out this bias because the transition probabilities were averaged arithmetically. 

In general, variational Bayes is also known to have a similar difficulty when tightly coupled variables are constrained to be independent in the variational approximation. Interestingly, in this experiment, it was able to attenuate the effect by making $\leftFactor$ and $\rightFactor$ small in magnitude, thereby reducing the coupling between them.

\begin{figure}
\includegraphics[width=\evalFigWSmall]{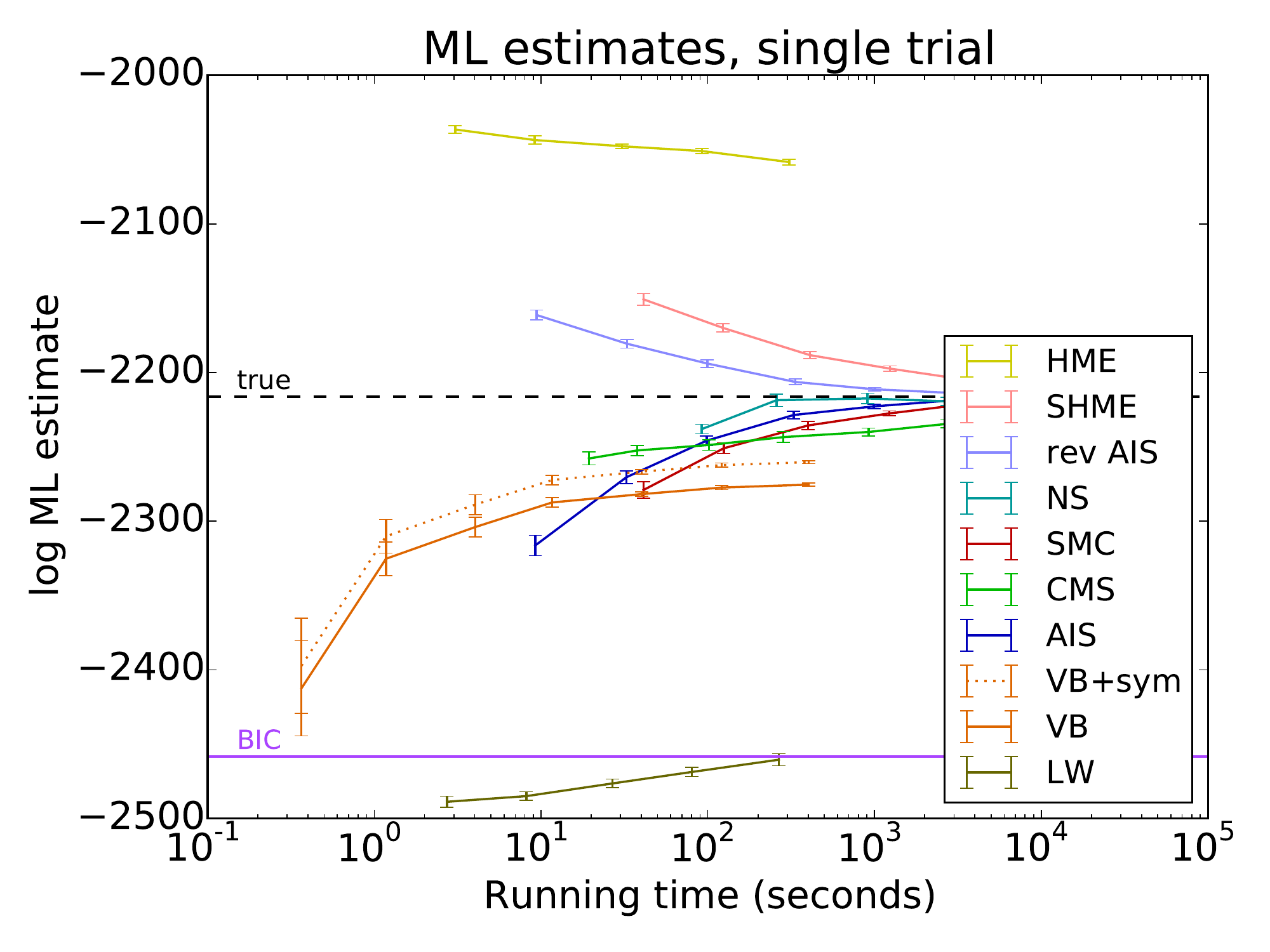}
\hspace{\evalFigSpSmall}
\includegraphics[width=\evalFigWSmall]{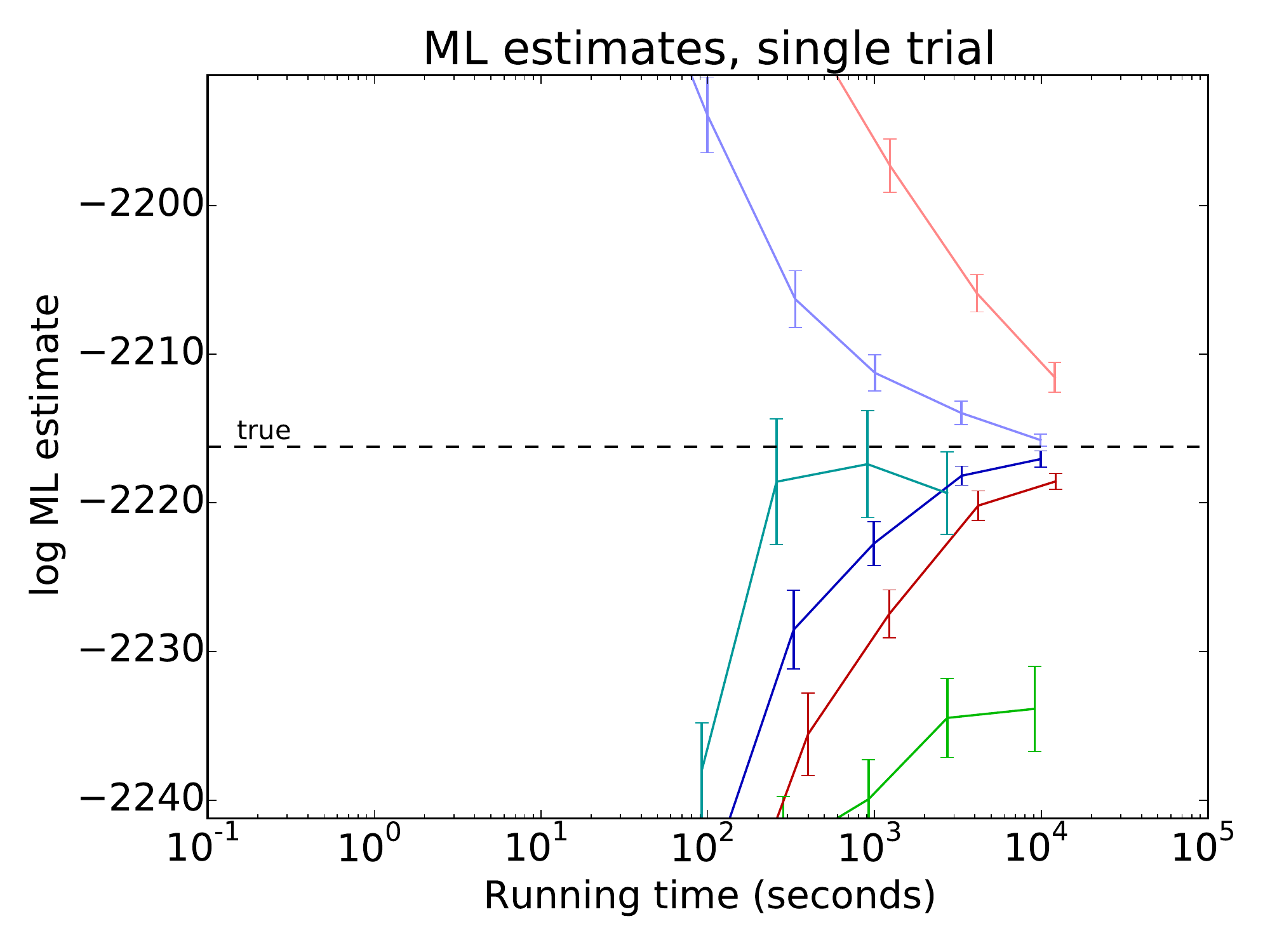}
\hspace{\evalFigSpSmall}
\includegraphics[width=\evalFigWSmall]{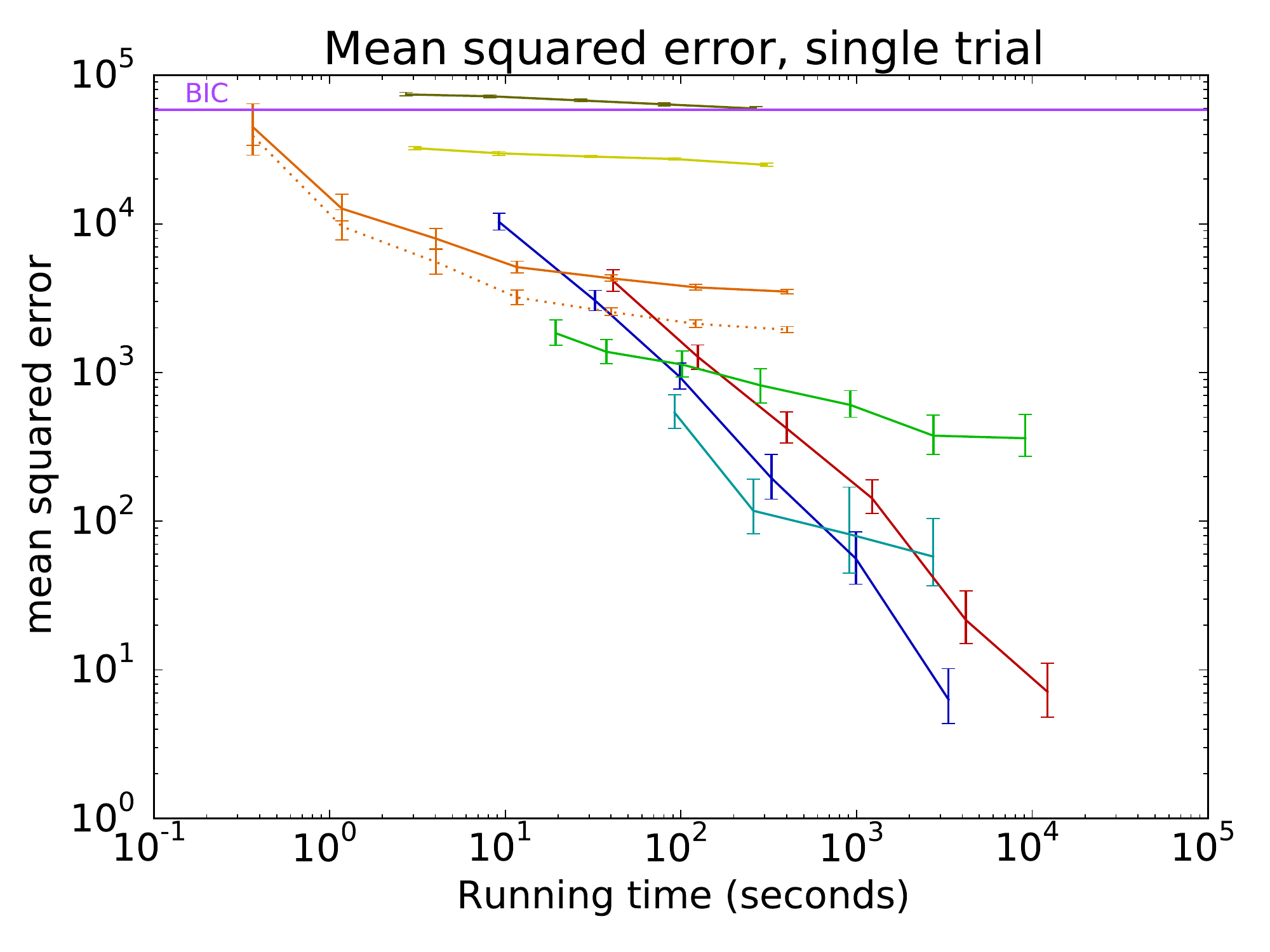}
\caption{Comparison of marginal likelihood estimators on the binary attribute model. {\bf Left:} average log-ML estimates across the 25 trials. {\bf Middle:} same as left, but zoomed in. {\bf Right:} MSE of individual samples. See Figure~\ref{fig:clustering-estimates} for the abbreviation key.}
\label{fig:binary-comparison}
\end{figure}

{\bf Binary attributes.} Finally, the results for the binary attribute model are shown in Figure~\ref{fig:binary-comparison}. Similarly to the clustering model, three algorithms came within 10 nats of the true value: NS, AIS, and SMC. NS and AIS each crossed the 10 nat threshold in similar amounts of time: AIS achieved an RMSE of 7.5 nats in 16.5 minutes, while NS achieved an RMSE of 9.0 nats in 15.1 minutes. By contrast, SMC achieved RMSE values of 11.9 and 4.7 in 20.5 minutes and 69 minutes, respectively. AIS and SMC continued to give more accurate results with increased computation time, while the accuracy of NS was hindered by the variance of the estimator. Overall, this experiment suggests that estimating the ML of a binary attribute model remains a difficult problem. 15 minutes is a very long time for a dataset with only 50 data points and 25 input dimensions.

\section{Discussion}

Based on our experiments, we observe that no single algorithm consistently dominated the others. The relative performance of different estimators varied tremendously depending on the model class, and it seems likely that other factors, such as the signal-to-noise ratio or the number of data points, could make a big difference. More work is required to understand which algorithms perform well on which models and why. However, we can draw some tentative recommendations from our results. As a general rule of thumb, we would suggest trying AIS first, because in all of our experiments, it achieved accurate results given enough intermediate distributions. If AIS is too slow, then SMC and NS are also worth considering. Likelihood weighting, the harmonic mean estimator, and the BIC are unlikely to give accurate results. These recommendations are consistent with the folklore in the field, and it is reassuring that we can now support them with quantitative evidence.

Interestingly, of the three strongest performing algorithms in our experiments---AIS, SMC, and NS---both AIS and SMC are instances of bridging between a tractable distribution and an intractable one using a sequence of intermediate distributions (see Section~\ref{sec:relationship-smc-ais}). Any algorithm which shares this structure can be reversed using our proposed technique to obtain a stochastic upper bound on the log-ML of simulated data. Therefore, if better algorithms are devised which build upon AIS and SMC (either separately or in combination), they can automatically be used in the context of BDMC to obtain more precise ground truth log-ML values on simulated data. 

We believe the ability to rigorously and quantitatively evaluate algorithms is what enables us to improve them. In many application areas of machine learning, especially supervised learning, benchmark datasets have spurred rapid progress in developing new algorithms and clever refinements to existing algorithms. One can select hyperparameters, such as learning rates or the number of units in a neural network, by quantitatively measuring performance on held-out validation data. This process is beginning to be automated through Bayesian optimization \citep{bayesian-optimization}. So far, the lack of quantitative performance evaluations in marginal likelihood estimation, and in sampling-based inference more generally, has left us fumbling around in the dark. ML estimators often involve design choices such as annealing schedules or which variables to collapse out. Just as careful choices of learning rates and nonlinearities can be crucial to the performance of a neural network, similar algorithmic hyperparameters and engineering choices may be crucial to building effective samplers and marginal likelihood estimators. We hope the framework we have presented for quantitatively evaluating ML estimators will enable the same sort of rapid progress in posterior inference and ML estimation which we have grown accustomed to in supervised learning.

\begin{small}
\bibliography{sandwich}
\end{small}

\appendix

\section{Caveats about marginal likelihood}
\label{app:ml-caveats}

While the focus of this work is on the algorithmic issues involved in estimating ML, we should mention several caveats concerning the ML criterion itself. First, the very notion of a ``correct'' or ``best'' model (as used to motivate Bayesian model comparison) may not be meaningful if none of the models under consideration accurately describe the data.
%\TODO{cite}. 
In such cases, different models may better capture different aspects of the data, and the best choice is often application-dependent. This caveat should be kept in mind when applying the methods of this paper: since the estimators are evaluated on simulated data, the results might not be representative of practical situations if the model is a poor match to the data. In general, ML should not be applied blindly, but should rather be used in conjunction with model checking methods such as posterior predictive checks \citep[chap.~6]{bayesian-data-analysis}.
%\TODO{cite}.

Another frequent criticism of ML is that it is overly sensitive to the choice of hyperparameters, such as the prior variance of the model parameters \citep{kass-ml-review, kass-raftery-ml-review}. Predictive criteria such as predictive likelihood and held-out error are insensitive to these hyperparameters in the big data setting, because with enough data, the likelihood function will overwhelm the prior. By contrast, the ML can be significantly hurt by a poor choice of hyperparameters, even for arbitrarily large datasets. This sensitivity to hyperparameters can lead to a significant bias towards overly simple models, since the more parameters a model has, the stronger the effect of a poorly chosen prior. We note, however, that this problem is not limited to the practice of explicitly comparing models by their ML: it also applies to the (more common) practice of tuning model complexity as part of posterior inference, for instance using reversible jump MCMC \citep{reversible-jump} or Bayesian nonparametrics \citep{ghahramani-bnp-tutorial}. Just as with explicit ML comparisons, these techniques can suffer from the bias towards simple models when the priors are misspecified.

Several techniques have been proposed which aim to alleviate the problem of hyperparameter sensitivity. \citet{intrinsic-bayes-factor} proposed the \emph{intrinsic
Bayes factor}, which is the probability of the data conditioned on a small number of data points. This can be equivalently viewed as computing the ratio of marginal likelihoods of different size datasets. Fractional Bayes factors \citep{fractional-bayes-factor} have a similar form, but the denominator includes all of the data points, and each likelihood term is raised to a power less than 1. Another approach maximizes the ML with respect to the hyperparameters; this is known as empirical Bayes, the evidence approximation, or type-II maximum likelihood \citep{mackay-hyper}. The motivation is that we can optimize over a small number of hyperparamters without overfitting too badly. Others suggest using ML, but designing the priors such that a poor choice of hyperparameters doesn't favor one model over another \citep{bde,neal-transfer}. We note that all of these alternative approaches require computing high-dimensional integrals over model parameters and possibly latent variables, and these integrals closely resemble the ones needed for ML. Therefore, one will run into the same computational obstables as in computing ML, and the techniques of this paper will still be relevant.

\section{Testing correctness of the implementation}
\label{app:testing}

ML estimators are notoriously difficult to implement correctly, for several reasons. First, an ML estimator returns a scalar value, and it's not clear how to recognize if that value is far off. This is in contrast with supervised learning, where one can spot of the algorithm is making silly predictions, or much unsupervised learning, where it is apparent when the algorithm fails to learn important structure in the data.
Furthermore, buggy MCMC transition operators often yield seemingly plausible posterior samples, yet lead to bogus ML estimates when used in an algorithm such as AIS. For these reasons, we believe ML estimation presents challenges for testing which are unusual in the field of machine learning. In this section, we discuss methods for testing mathematical correctness of ML estimator implementations.

In this work, we used several strategies, which we recommend following in any work involving ML estimation. \citet{testing-mcmc} discuss some of these techniques in more detail.
\begin{enumerate}
\item Most of the MCMC operators were implemented in terms of functions which returned conditional probability distributions. (The distributions were represented as classes which knew how to sample from themeslves and evaluate their density functions.) The conditional probability computations can be ``unit tested'' by checking that they are consistent with the joint probability distribution. In particular,
\[ \frac{\pmf(x \given u)}{\pmf(x^\prime \given u)} = \frac{\pmf(x, u)}{\pmf(x^\prime, u)} \]
must hold for \emph{any} triple $(x, x^\prime, u)$. This form of unit testing is preferable to simulation-based tests, because the identity must hold exactly, and fails with high probability if the functions computing conditional probabilities are incorrect.

Analogously, the updates for variational Bayes were tested by checking that they returned local maxima of the variational lower bound. 

\item To test the MCMC algorithms themselves, we used the Geweke test \citep{geweke-test}. This can be thought of as an ``integration test,'' since it checks that all of the components of the sampler are working together correctly. This test is based on the fact that there are two different ways of sampling from the joint distribution over parameters $\params$, latent variables $\latent$, and data $\obs$. First, one can sample forwards from the model. Second, one can begin with a forwards sample and alternate between (a) applying the MCMC transition operator, which preserves the posterior $\pmf(\params, \latent \given \obs)$, and (b) resampling $\obs$ from $\pmf(\obs \given \params, \latent)$. If the implementation is correct, these two procedures should yield samples from \emph{exactly} the same distribution. One can check this by checking P-P plots of various statistics of the data. 

The Geweke test is considered the gold standard for testing MCMC algorithms. It can detect surprisingly subtle bugs, because the process of resampling the data tends to amplify small biases in the sampler. (E.g., if the MCMC operator slightly overestimates the noise, the data will be regenerated with a larger noise; this bias will be amplified over many iterations.) The drawback of the Geweke test is that it gives no indication of where the bug is. Therefore, we recommend that it be run only after all of the unit tests pass.

\item The ML estimators were tested on toy distributions, where the ML could be computed analytically, and on very small instances of the clustering and binary models, where it could be computed through brute force enumeration of all latent variable configurations.

\item Because we had implemented a variety of ML estimators, we could check that they agreed with each other on easy problem instances: in particular, instances with extremely small or large single-to-noise ratios (SNR), or small numbers of data points. 
\end{enumerate}

The vast majority of bugs that we caught were caught in step 1, only a handful in steps 2 and 3, and none in step 4. We would recommend using 1, 2, and 3 for any work which depends on ML estimation. (Steps 1 and 3 are applicable to partition function estimation more generally, while the Geweke test is specific to directed models.) Step 4 may be overkill for most applications because it requires implementing multiple estimators, but it provides an additional degree of reassurance in the correctness of the implementation.

The techniques of this section test only the \emph{mathematical correctness} of the implementation, and do not guarantee that the algorithm returns an accurate answer. The algorithm may still return inaccurate results because the MCMC sampler fails to mix or because of statistical variability in the estimator. These are the effects that the experiments of this paper are intended to measure.

\end{document}